%% file: main.tex
\documentclass[lettersize,journal]{IEEEtran}
\IEEEoverridecommandlockouts      
\usepackage[dvipsnames,table,xcdraw]{xcolor}
\usepackage{graphics} 
\usepackage{graphicx} %
\usepackage{lipsum}
\usepackage{tikz}
\usetikzlibrary{positioning, shapes.geometric}
\usepackage{verbatim}
\usetikzlibrary{calc}
\usepackage[font=footnotesize]{caption}
\usepackage{wrapfig}

\usepackage{amsmath} 
\usepackage{amssymb}  
\usepackage{mathtools}

\usepackage{cite}

\usepackage{enumitem}
\usepackage{mleftright}
\usepackage{balance}

\graphicspath{ {../figs/}{../figures/}}

\title{\LARGE \bf Image-Based Trajectory Tracking through Unknown Environments
without Absolute Positioning}

\author{{Shiyu Feng$^{1,\dagger}$, Zixuan Wu$^{2,\dagger}$, Yipu Zhao$^{3}$ and Patricio A. Vela$^{2}$}
\thanks{*This work supported in part by NSF Award \#1849333.}
\thanks{$\dagger$ Equal contribution}
\thanks{$^{1}$S. Feng is with the School of Mechanical Engineering and the School of Electrical and Computer Engineering, Georgia Institute of Technology, Atlanta, GA 30308, USA.
		{\tt\small shiyufeng@gatech.edu}}
\thanks{$^{2}$Z. Wu and P.A. Vela are with the School of Electrical and Computer Engineering and the Institute for Robotics and Intelligent Machines, Georgia Institute of Technology, Atlanta, GA 30308, USA.
        {\tt\small \{zwu380, pvela\}@gatech.edu}}%
\thanks{$^{3}$Y. Zhao is with Meta Reality Lab, Redmond, USA. {\tt\small zhaoyipu@gmail.com}}
}

\begin{document}

\maketitle
\thispagestyle{plain}
\pagestyle{plain}

\begin{abstract}
This paper describes a stereo image-based visual servoing system for
trajectory tracking by a non-holonomic robot without externally derived 
pose information nor a known visual map of the environment.  It is
called {\em trajectory servoing}. The critical component is a feature-based, indirect Simultaneous Localization And Mapping (SLAM) method to
provide a pool of available features with estimated depth, so that they
may be propagated forward in time to generate image feature
trajectories for visual servoing. Short and long distance experiments
show the benefits of trajectory servoing for navigating unknown areas
without absolute positioning.  Empirically, trajectory servoing
has better trajectory tracking performance than pose-based
feedback when both rely on the same underlying SLAM system. 
\end{abstract}


\input{intro.tex}

\input{background.tex}

\input{contrib.tex}
\input{review.tex}

\input{shortDist.tex}

\input{shortExp.tex}
\input{longDist.tex}

\input{longExp.tex}
\input{conc.tex}




\appendices
\input{append.tex}

\bibliographystyle{IEEEtran}

\bibliography{teachandrepeat,visualservo,priorwork,SLAMNav}



%
\begin{IEEEbiography}[{\includegraphics[width=1in,height=1.25in,clip,keepaspectratio]{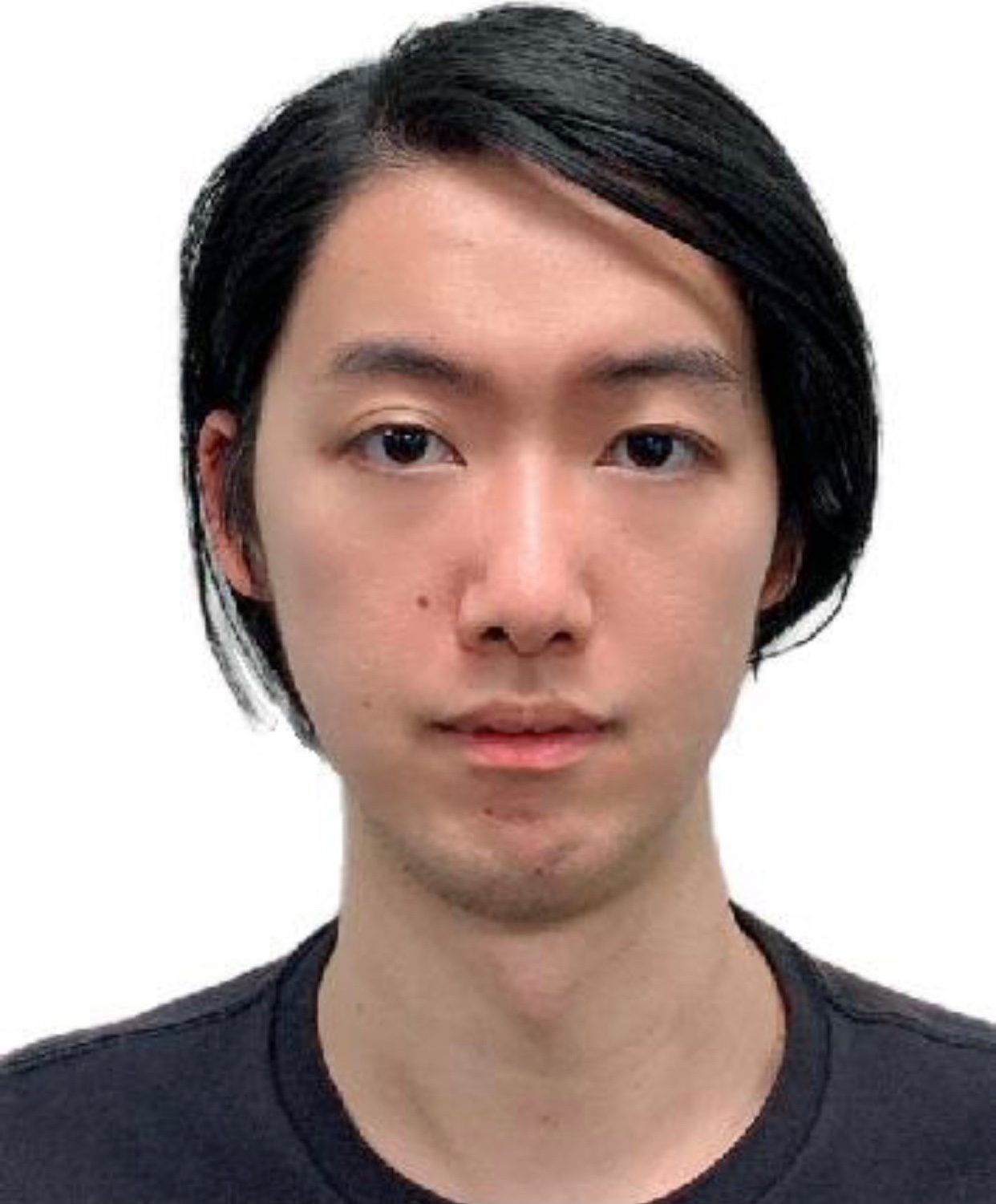}}]{Shiyu Feng} 
is a Ph.D. candidate in the School of Mechanical Engineering, at the Georgia Institute of Technology. 
He is a member of Intelligent Vision and Automation Laboratory 
(IVALab) supervised by Dr. Patricio A. Vela.
His research focuses on vision-based hierarchical navigation 
using sparse representation to improve the computational efficiency and 
scalability.
Previously, he earned his B.Eng and M.Eng in Mechanical Engineering from 
Chongqing University, China (2015) and University of California, Berkeley (2016).

Shiyu Feng is a student member of IEEE.
\end{IEEEbiography}

\vskip -2\baselineskip plus -1fil

\begin{IEEEbiography}[{\includegraphics[width=1in,height=1.25in,clip,keepaspectratio]{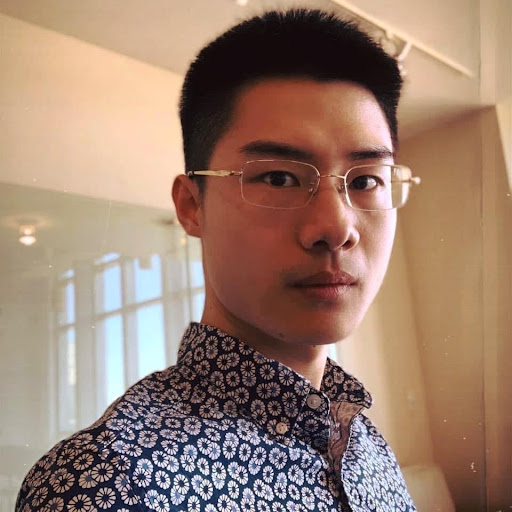}}]{Zixuan Wu}
received his B.Eng. degree in automation from Harbin Institute of Technology, 
China in 2019. After that, he went to Georgia Institute of Technology, USA, 
obtained his M.Sc. degree at School of Electrical and Computer Engineering, 
and started the Ph.D. at the same department in 2021. 
His research interests lie in reinforcement learning, visual servoing 
and visual navigation. 
Recently, he works on the experience replay optimization for heterogeneous robot teaming.
\end{IEEEbiography}

\vskip -2\baselineskip plus -1fil

\begin{IEEEbiography}[{\includegraphics[width=1in,height=1.25in,clip,keepaspectratio]{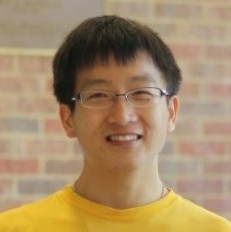}}]{Yipu Zhao}
is a Research Scientist at Meta Reality Lab. 
Prior to joining Meta, he obtained his Ph.D. in 2019, 
under the supervision of Patricio A. Vela, at the School of Electrical and 
Computer Engineering, Georgia Institute of Technology, USA.  
Previously he received his B.Sc. degree in 2010 and M.Sc. degree in 2013, 
at the Institute of Artificial Intelligence, Peking University, China.  
His research interests include visual odometry/SLAM, 3D reconstruction, 
and multi-object tracking.

Dr. Zhao is a member of IEEE.
\end{IEEEbiography}

\vskip -2\baselineskip plus -1fil

\begin{IEEEbiography}[{\includegraphics[width=1in,height=1.25in,clip,keepaspectratio]{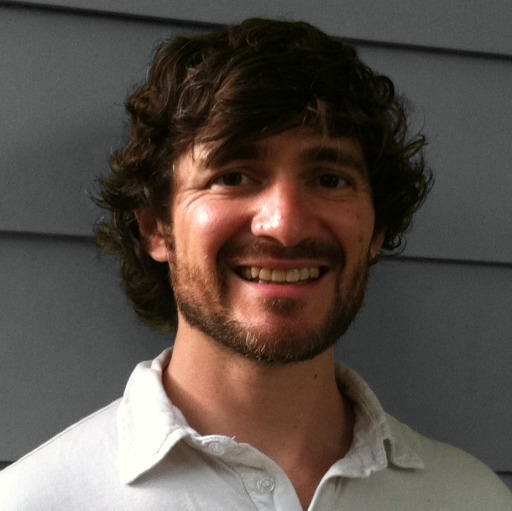}}]{Patricio A. Vela}
is an associate professor in the School of Electrical and Computer Engineering, and the Institute of Robotics and Intelligent Machines, at Georgia Institute of Technology, USA. His research interests lie in the geometric perspectives to control theory and computer vision. Recently, he has been interested in the role that computer vision can play for achieving control-theoretic objectives of (semi-)autonomous systems. His research also covers control of nonlinear systems, typically robotic systems.

Prof. Vela earned his B.Sc. degree in 1998 and his Ph.D. degree in control and dynamical systems in 2003, both from the California Institute of Technology, where he did his graduate research on geometric nonlinear control and robotics. In 2004, Dr. Vela was as a post-doctoral researcher on computer vision with School of ECE, Georgia Tech. He join the ECE faculty at Georgia Tech in 2005.

Prof. Vela is a member of IEEE.
\end{IEEEbiography}

\end{document}

%% file: intro.tex
\section{Introduction \label{sec:Intro}}
Navigation systems with real-time needs often employ hierarchical schemes
that decompose navigation across multiple spatial and temporal scales.
Doing so improves real-time responsiveness to novel information gained
from sensors, while being guided by the more slowly evolving global
path. At the lowest level of the hierarchy lies trajectory tracking to
realize the planned paths or synthesized trajectories. In the absence of
an absolute reference (such as GPS) and of an accurate map of the
environment, only on-board mechanisms can support trajectory-tracking.
These include odometry through {\em proprioceptive} sensors (wheel
encoders, IMUs, etc.) or visual sensors.  Pose estimation from
proprioceptive sensors is not observable, thus visual sensors provide
the best mechanism to anchor the robot's pose estimate to external,
static reference locations.  Indeed visual odometry (VO) or visual SLAM
(V-SLAM) solutions are essential in these circumstances.
However, VO/V-SLAM estimation error and pose drift
degrade trajectory tracking performance. Is it possible to do better?

The hypothesis explored in this paper is that {\em performing trajectory
tracking in the image domain reduces the trajectory
tracking error of systems reliant on VO or V-SLAM pose estimates for feedback}.
Trajectory tracking feedback shifts from pose space to perception space.  
Perception space approaches have several favorable properties when used
for navigation \cite{Smith2017,Smith2020}.  Shifting the representation
from being world-centric to being viewer-centric reduces computational
demands and improves run-time properties.
For trajectory tracking without reliable absolute pose information,
simplifying the feedback pathway by skipping processes that
degrade performance of--or are not essential to--the local
tracking task may have positive benefits. 
Using image measurements to control motion relative to visual
landmarks is known as {\em visual servoing}. Thus, the objective
is to explore the use of image-based visual servoing for long
distance trajectory tracking with a stereo camera as the primary
sensor and {without absolute positioning}.  The technique, which we
call {\em trajectory servoing}, will be shown to
improve trajectory tracking 
over systems reliant on V-SLAM for pose-based feedback.  {In this
paper, a trajectory is a time parametrized curve in Cartesian space.}

%% file: background.tex
\subsection{Related Work \label{sec:BG}}

\subsubsection{Visual Servoing\label{sec:BG_vs}}
Visual servoing (VS) has a rich history and a diverse set of strategies for
stabilizing a camera to a target pose described visually. VS algorithms
fit into one of two categories:
image-based visual servoing (IBVS) and 
position-based visual servoing (PBVS)\cite{4015997,4141039}. 
IBVS implementations include both feature stabilization and feature
trajectory tracking \cite{4141039,6564708}. 
IBVS emphasizes point-to-point reconfiguration given a
terminal image state \cite{virtualCam}. It requires artificial markers
and co-visibility of image measurements during motion
\cite{visibilitySol}, which may not be satisfied for long-distance
displacements. Furthermore, there is no guarantee on the path taken 
since the feature space trajectory has a nonlinear relationship with the
Cartesian space trajectory.
Identifying a feature path to track based on a target Cartesian space
trajectory requires mapping the target positions into the image frame
over time to generate the feature trajectory \cite{6564708}.

The target application is trajectory tracking for a mobile robot {with IBVS}.
{Both holonomic \cite{1642112} and non-holonomic \cite{SEGVIC2009172,inproceedingsNwL,cherubini:hal-00639659,9197114,5740604,1570628}} 
mobile robots have been studied as candidates for visual servoing.
Most approaches do not use the full IBVS equations involving the image
Jacobian.  The centroid of the features \cite{SEGVIC2009172,5740604},
the most frequent horizontal displacement of the matched feature pairs
\cite{inproceedingsNwL} or other qualitative cost functions
\cite{1642112} are used to generate \cite{SEGVIC2009172} or correct
\cite{inproceedingsNwL} the feedforward angular velocity of mobile
robot. 
These simplifications work well in circumstances tolerant to
high tracking inaccuracy (e.g. an outdoor, open field navigation).
In this paper, we use more precise velocity relations between the robot
and feature motion to generate a feedback control signal for exact
tracking similar to \cite{cherubini:hal-00639659}. 
That work studied the path following problem with a visible path marker
line, which does not hold here.




\subsubsection{Visual Teach and Repeat\label{sec:BG_vtr}}

Evidence that visual features can support trajectory tracking or
consistent navigation through space lies in the {\em Visual Teach and
Repeat} (VTR) navigation problem in robotics 
\cite{SEGVIC2009172,inproceedingsNwL}. 
Given data or recordings of prior paths through an environment, robots
can reliably retrace past trajectories.
The teaching phase of VTR creates a visual map that contains features
associated with robot poses obtained from visual odometry 
\cite{SEGVIC2009172,8967994,High,longrange,KRAJNIK2017127}. 
Extensions include real-time construction of the VTR data structure during the teaching process, and the maintenance and updating of the VTR data during repeat runs \cite{High,8967994}.  
Feature descriptor improvements make the feature matching more robust to
the environment changes \cite{KRAJNIK2017127,8460674}.
Visual data in the form of feature points can have task relevant and irrelevant features, which provide VTR algorithms an opportunity to select a subset that best contributes to the localization or path following task \cite{8967994,longrange}. 
It is difficult to construct or update a visual map in
real-time while in motion due to the separation of the teach and repeat
phases. In addition, VTR focuses more on local map consistency and does
not work toward global pose estimation \cite{longrange} compared with
SLAM since the navigation problems it solves are usually defined in the
local frame.

Another type of VTR uses the optical flow
\cite{inproceedingsNwL,10.1007/s10846-015-0320-1} or feature
sequences \cite{KIDONO2002121,6816849,5649557} along the trajectory,
which is then encoded into a VTR data structure and control algorithm
in the teaching phase.  Although similar to visual servoing, the system
is largely over-determined.  It can tolerate feature tracking failure,
compared with traditional visual servo system, but may lead to
discontinuities \cite{1389589}.  Though this method handles long
trajectories, and may be supplemented from new teach recordings, 
it can only track paths through visited space.

\subsubsection{Navigation Using Visual SLAM\label{sec:BG_slam}}

Visual Simultaneous Localization And Mapping (V-SLAM) systems estimate
the robot's trajectory and world structure as the robot moves
through space \cite{cadena2016past}.  
SLAM derived pose estimation naturally supports navigation or path
following \cite{fontanel, hybrid}, and PBVS \cite{structure, gap}. Most of these methods
still need to initialize and maintain topological or metric visual maps
for path planning \cite{fontanel, hybrid} or for localizing the robot for PBVS \cite{structure, gap}.
These works do not involve Cartesian trajectories \cite{fontanel,
structure, gap} nor provide solutions to occlusion and tracking loss problems \cite{fontanel, hybrid, structure, gap}. 
The shortcomings imply the inability to track long Cartesian
trajectories through unknown environments.

Pose estimation accuracy of SLAM is a major area of study
\cite{8964284,7140009,8436423,8794369,8460664}. 
However, most studies only test under open-loop conditions \cite{8964284,7140009,8436423,8460664}, 
i.e., they analyze the pose estimation difference with the ground
truth trajectory and do not consider the error induced when the
estimated pose informs feedback control. 
More recently, closed-loop evaluation of V-SLAM algorithms as part of
trajectory tracking feedback control or navigation were tested for
individual SLAM systems
\cite{cvivsic2018soft,WeEtAl_VIOSwarm[2018],lin2018autonomous} or across
different systems \cite{zhao2020closednav}.  
The studies exposed the influence of V-SLAM estimation
drift and latency on pose-based feedback control performance.  This
paper builds upon an existing closed-loop benchmarking framework
\cite{zhao2020closednav} and shows improved tracking performance by
trajectory servoing.

%% file: contrib.tex
\subsection{Contribution \label{sec:BG_cont}}
Both IBVS and VTR depend on reliably tracked, known
features within the field-of-view, which can only be achieved by 
artificial, task-oriented scenarios \cite{6564708,cherubini:hal-00639659} or 
a pre-built trajectory map \cite{SEGVIC2009172,8967994,High,longrange,KRAJNIK2017127}.
This limitation does not permit travel through unknown environments to
an unseen terminal state, and motivates the use of feature-based V-SLAM
systems with online map saving \cite{fontanel,hybrid}. 
Compared with PBVS based on SLAM \cite{structure, gap}, trajectory
servoing uses IBVS to less frequently query the pose from a stereo
V-SLAM system \cite{8964284}, thereby attenuating the impact of
estimation drift and error on trajectory tracking performance.

Trajectory servoing bypasses the explicit use of VO/V-SLAM pose, 
whose estimates are vulnerable to image-driven uncertainty which
manifests as pose error or drift \cite{4015997,4141039},
by relying on the V-SLAM feature maintenance components that
provide accurate and robust feature tracking and mapping. 
We present evidence for the assertion that the coupling between V-SLAM
and IBVS combines their advantages to provide more effective feedback
signals relative to V-SLAM pose-based control. 
Simulation and real experiment performance benchmarking show that trajectory
servoing improves trajectory tracking performance
over pose-based feedback using SLAM estimates, thereby mitigating
the effect of pose estimation error or drift, even though the same visual
information is used for closing the loop.
The beneficial coupling is the motivation behind the trajectory servoing
system design. It is a promising approach to trajectory tracking through
unknown environments in the absence of absolute positioning signals.

%% file: review.tex
\subsection{Image-Based Visual Servoing Rate Equations\label{sec:ibvs}}

The core algorithm builds on IBVS \cite{538972}. 
This section covers IBVS with an emphasis on how it relates the
velocity of image features to the robot velocities via the image Jacobian
\cite{4015997,4141039}. These equations will inform the trajectory
tracking problem under non-holonomic robot motion. We use the more
modern notation from geometric mechanics \cite{MLS} since it provides 
equations that better connect to contemporary geometric control
and to SLAM formulations for moving rigid bodies.

\newcommand{\Real}{\mathbb{R}}
\newcommand{\set}[1]{\left\{\, #1 \,\right\}}
\newcommand{\of}[1]{\left( #1 \right)}
\newcommand{\inverse}[1]{#1^{-1}}   
\newcommand{\mc}[1]{{\mathcal{#1}}}
\newcommand{\mcframe}[3]{{#1}^{\mc{#2}}_{\mc{#3}}}
\newcommand{\LA}[1]{\mathfrak{#1}}
\newcommand{\Ad}{{\textrm{Ad}}}

\newcommand{\roboVel}{\boldsymbol{u}}
\newcommand{\roboVelDes}{\boldsymbol{u}^{*}}
\newcommand{\roboF}{g}
\newcommand{\roboBV}{\xi_{\roboVel}}
\newcommand{\camF}{h}
\newcommand{\camBV}{\zeta_{\roboVel}}
\newcommand{\camProj}{\boldsymbol{H}}
\newcommand{\camPt}{r}
\newcommand{\funcToMat}{\boldsymbol{M}}

\newcommand{\imJac}{\boldsymbol{\mathcal{L}}}
\newcommand{\imJacV}{\boldsymbol{L}}

\newcommand{\geoJac}{{G}}

\newcommand{\numFeat}{n_{\text{F}}}

\newcommand{\pose}{\boldsymbol{P}}
\newcommand{\control}{\boldsymbol{u}}
\newcommand{\transCR}{^C\boldsymbol{T}_{R}}
\newcommand{\transCRv}{^C\boldsymbol{T}_{R,v}}
\newcommand{\transCRw}{^C\boldsymbol{T}_{R,\omega}}

\newcommand{\pointSet}{\boldsymbol{Q}}
\newcommand{\pointSetV}{\boldsymbol{q}}
\newcommand{\pointIndividual}{{q}}

\newcommand{\featureSet}{\boldsymbol{S}}
\newcommand{\featSetV}{\boldsymbol{s}}
\newcommand{\featSetDV}{\boldsymbol{s}^*}
\newcommand{\featSetDVdot}{\dot{\boldsymbol{s}}^*}
\newcommand{\featureSetA}{\boldsymbol{S}_a}
\newcommand{\featureSetD}{\boldsymbol{S}^*}
\newcommand{\featureSetDA}{\boldsymbol{S}_a^*}

\newcommand{\error}{\boldsymbol{e}}
\newcommand{\errorSet}{\boldsymbol{E}}

\newcommand{\imageJacobian}{\boldsymbol{L}}

\newcommand{\imageJacobiani}{\boldsymbol{L}_{S_i}}
\newcommand{\imageJacobianv}{\boldsymbol{L}_{S,v}}
\newcommand{\imageJacobianw}{\boldsymbol{L}_{S,\omega}}
\newcommand{\transFeature}{g}
\newcommand{\homo}{H}



\subsubsection{Non-Holonomic Robot and Camera Kinematic Models}

Let the motion model of the robot be a kinematic Hilare robot model
where the pose state
$\mcframe{\roboF}{W}{R} \in SE(2)$ evolves under the control $\roboVel = [\nu,
\omega]^T$ as
\begin{equation} \label{eq:roboDE}
  \mcframe{\dot \roboF}{W}{R} 
    = \mcframe{\roboF}{W}{R} 
      \cdot \begin{bmatrix} 1 & 0 \\ 0 & 0 \\ 0 & 1 \end{bmatrix}
      \begin{bmatrix} \nu \\ \omega \end{bmatrix}
    = \mcframe{\roboF}{W}{R} \cdot \roboBV,
\end{equation}
for $\nu$ a forward linear velocity and $\omega$ an angular velocity, and 
$\roboBV \in \LA{se}(2)$.
The state is the robot frame $\mc{R}$ relative to the world frame
$\mc{W}$. The camera frame $\mc{C}$ is presumed to be described as 
$\mcframe{\camF}{R}{C}$ relative to the robot frame. Consequently, camera
kinematics relative to the world frame are
\begin{equation} \label{eq:camDE}
  \mcframe{\dot \camF}{W}{C} 
    = \mcframe{\roboF}{W}{R} \cdot \mcframe{\camF}{R}{C}
      \cdot \Ad^{-1}_{\mcframe{h}{R}{C}} \cdot 
      \begin{bmatrix} 1 & 0 \\ 0 & 0 \\ 0 & 1 \end{bmatrix}
      \begin{bmatrix} \nu \\ \omega \end{bmatrix}
    = \mcframe{\roboF}{W}{R} \cdot \mcframe{\camF}{R}{C}
      \cdot \camBV,
\end{equation}
with $\camBV \in \LA{se}(2)$.
Now, let the camera projection equations be given by the function
$\camProj:\Real^3 \rightarrow \Real^2$ such that a point $\pointIndividual^{\mc{W}}$
projects to the camera point $\camPt = \camProj \circ \mcframe{\camF}{C}{W}
(\pointIndividual^{\mc W})$. Under camera motion, the differential
equation relating the projected point to the camera velocity for a
static point $\pointIndividual^{\mc W}$ is
\begin{equation} \label{eq:camDEsimp}
  \dot \camPt = \text{D}\camProj(\pointIndividual^{\mc C}) \cdot
    \left( \camBV \cdot \pointIndividual^{\mc C} \right),
    \quad \text{for} \ 
    \pointIndividual^{\mc C} = \mcframe{\camF}{C}{W} \pointIndividual^{\mc W},
\end{equation}
where 
$\text{D}$ is the differential operator. 
Since the operation 
$\zeta \cdot \pointIndividual$ is linear for $\zeta \in \LA{se}(2)$, $\pointIndividual \in \Real^3$, it can
be written as a matrix-vector product $\funcToMat(\pointIndividual) \zeta$ leading to,
\begin{equation} \label{eq:camPtDot}
  \dot \camPt = \text{D}\camProj(\pointIndividual^{\mc C}) \cdot \funcToMat(\pointIndividual^{\mc C}) \camBV 
    = \imJac(\pointIndividual^{\mc C}) \camBV,
\end{equation}
where $\imJac:\Real^3 \times \LA{se}(2)$ is the Image Jacobian. Given the point and projection pair
$(\pointIndividual,\camPt) \in \Real^3 \times \Real^2$, $\imJac$ works out to be
\begin{equation} \label{eq:imJac}
  \imJac(\pointIndividual) = \imJac(\pointIndividual,\camPt) = 
    \begin{bmatrix*}
      - \frac{f}{\pointIndividual^3} & 0 
      & \camPt^2 \\[1pt]
      0 & -\frac{f}{\pointIndividual^3} 
      & -\camPt^1
    \end{bmatrix*},
\end{equation}
where $f$ is the focal length. 
Recall that $\camPt = \camProj(\pointIndividual)$. Re-expressing
$\imJac$ as a function of $(\pointIndividual,\camPt)$
simplifies its written form, and exposes what information is available
from the image directly $\camPt \in \Real^2$ and what additional information
must be known to compute it: coordinate $\pointIndividual^3$ from $\pointIndividual^{\mc C} \in
\Real^3$ in the camera frame, which is also called depth. 
With a stereo camera, the depth value is triangulated. The next section
will use these equations for trajectory tracking with image features.

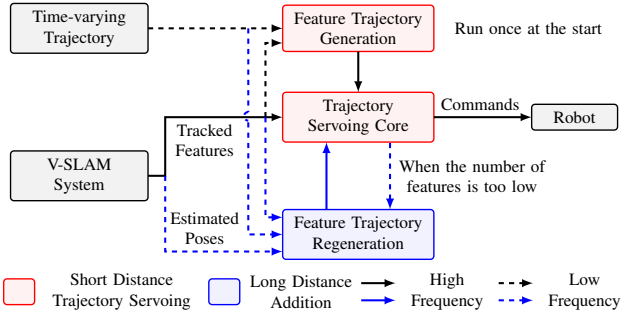
\begin{figure}[t]
  \vspace*{0.1in}
  \hspace*{-0.2in}
  \centering
  \scalebox{0.65}{\input{figs/flow_chart.tex}}
  \caption{A trajectory servoing system has two major components. 
  One (red) steers the robot to track short paths, while the other (blue)
  ensures the sufficiency of features to use by querying a V-SLAM module.
  The entire system is used when tracking long distance trajectories.
  Solid arrows indicate high frequency data passing, and 
  dashed arrows low frequency. All blue arrows represent the information 
  flow related to long distance addition.
  \label{fig:system_flowchart}}
  \vspace*{-0.2in}
\end{figure}

%% file: figs/flow_chart.tex
\tikzstyle{block} = [draw, rectangle, text centered, thick,rounded corners=2pt,
                     minimum height=1.5em, minimum width=5em, inner sep=4pt]
\tikzstyle{typical} = [fill=white!95!black]
\tikzstyle{reddish} = [draw=red,fill=white!95!red]
\tikzstyle{blueish} = [draw=blue,fill=white!95!blue]
\tikzstyle{greenish} = [draw=green!40!gray,fill=white!95!green]
\tikzstyle{longblock} = [draw,rectangle,text centered,thick,rounded corners=2pt,
                     minimum height=1.5em, minimum width=8em, inner sep=4pt]
\tikzstyle{largeBlock} = [draw, rectangle, very thick,
                     minimum height=19.3em, minimum width=25em, inner sep=4pt]
\tikzstyle{smallBlock} = [draw, rectangle, text centered, thick, dashed,
                     minimum height=13.25em, minimum width=15em, inner sep=4pt]
\tikzstyle{dashedBlock} = [draw, dashed, rectangle,
                     minimum height=2em, minimum width=4em, inner sep=4pt]
\tikzstyle{dottedBlock} = [draw, rectangle, text centered, ultra thick,
					 minimum height=1.5em, 
					 minimum width=8em, inner sep=4pt,
					 dash pattern=on 1pt off 2pt on 4pt off 2pt] 
\tikzstyle{newtip} = [->, very thick]
\tikzstyle{bidir} = [<->, very thick]
\tikzstyle{newtip_dashed} = [->, very thick, dashed]
\begin{tikzpicture}[auto, inner sep=0pt, outer sep=0pt, >=latex]

  \node[longblock, typical, text width=6em] (slam) {\centering V-SLAM \\ System};
  
  \node[longblock, typical, anchor=south, text width=6em] (traj) 
  at ($(slam.north) + (0, 2)$) {\centering Time-varying \\ Trajectory};
    
  \node[longblock,reddish,anchor=center, text width=8em] (feat_traj) 
    at ($(traj.east) + (4.3, 0)$) 
    {\centering Feature Trajectory \\ Generation};
  \node[longblock, reddish, anchor=center, text width=8em] (core)  
    at ($(slam.east)+(4.3, 1.2)$)
    {\centering Trajectory \\ Servoing Core};
  \node[longblock, blueish, anchor=center, text width=8em] (reg)  
    at ($(core.center)+(0, -2.4)$)
    {\centering Feature Trajectory \\ Regeneration};

  \node[block, typical, anchor=west] (robot) 
  at ($(core.east) + (2, 0)$) {\centering Robot};
  
  \node[anchor=south west, text centered, text width=9em] (reg_text) 
  at ($(reg.north east) + (-0.8, 0.35)$) 
  {\centering When the number of \\ features is too low};
  
  \node[anchor=west, text centered, text width=9em] (feat_text) 
  at ($(feat_traj.east) + (1em, 0)$) 
  {\centering Run once at the start};

  \draw[newtip] (slam.east) -- ($(slam.east)+(10pt,0pt)$) --
   ($(slam.east)+(10pt,1)$) |-
  node[below,anchor=north,text centered,text width=5em,xshift=23pt,yshift=-4pt]{Tracked \\  Features}(core.west);  
  \draw[newtip,dashed] ($(core.west)+(-10pt,0)$) --
   ($(feat_traj.west)+(-10pt,-0.3)$) -- ($(feat_traj.west)+(0,-0.3)$);
  \draw[newtip,dashed,draw=blue] ($(core.west)+(-10pt,0)$) --
   ($(reg.west)+(-10pt,0.35)$) -- ($(reg.west)+(0,0.35)$);
   
  \draw[newtip,dashed,draw=blue] ($(slam.east)+(10pt,0pt)$) --
   ($(slam.east)+(10pt,-0.8)$) |-
  node[above,anchor=south,text centered,text width=5em,xshift=23pt,yshift=4pt]{Estimated \\ Poses}($(reg.west)+(0,-0.35)$);  
  
  \draw[newtip,dashed] (traj.east) -- (feat_traj.west);
  \draw[-,very thick,dashed,draw=blue] ($(feat_traj.west)+(-20pt,0)$) -- ($(core.west)+(-20pt,0.3em)$);
  \draw[very thick, dashed,draw=blue] ($(core.west)+(-20pt,0.3em)$) arc (90:270:0.25em);
  \draw[newtip,dashed,draw=blue] ($(core.west)+(-20pt,-0.3em)$) --
   ($(reg.west)+(-20pt,0)$) -- ($(reg.west)+(0,0)$);
  \draw[newtip] (feat_traj.south) -- (core.north);
  
  \draw[newtip] (core.east) |-
  node[above,anchor=south,xshift=27pt,yshift=4pt]{Commands}(robot.west);
  
  \draw[newtip,draw=blue] ($(reg.north) + (-0.65,0)$) -- ($(core.south) + (-0.65,0)$);
  \draw[newtip,dashed,draw=blue,<-] ($(reg.north) + (0.65,0)$) -- ($(core.south) + (0.65,0)$);

  \node[longblock,reddish,anchor=north,minimum width=1em,text width=1em] (legend1) at ($(slam.south) + (-3.5em, -1.6)$) {};
  \node[anchor=west,text centered,text width=9em] (legend1_name) at ($(legend1.east) + (0.2, 0)$) {\centering Short Distance \\ Trajectory Servoing};
  
  \node[longblock,blueish,anchor=west,minimum width=1em,text width=1em] (legend2) at ($(legend1_name.east) + (0.2,0)$) {};
  \node[anchor=west,text centered,text width=6em] (legend2_name) at ($(legend2.east) + (0.2, 0)$) {\centering Long Distance \\ Addition};  
  
  \draw[newtip] ($(legend2_name.east) + (0.5em, 0.5em)$) -- ($(legend2_name.east) + (2.5em, 0.5em)$);
  \draw[newtip,draw=blue] ($(legend2_name.east) + (0.5em, -0.5em)$) -- ($(legend2_name.east) + (2.5em, -0.5em)$);
  \node[anchor=west,text centered,text width=5em] (legend3_name) at ($(legend2_name.east) + (3em, 0)$) {\centering High \\ Frequency};
  
  \draw[newtip,dashed] ($(legend3_name.east) + (0.5em, 0.5em)$) -- ($(legend3_name.east) + (2.5em, 0.5em)$);
  \draw[newtip,dashed,draw=blue] ($(legend3_name.east) + (0.5em, -0.5em)$) -- ($(legend3_name.east) + (2.5em, -0.5em)$);
  \node[anchor=west,text centered,text width=5em] (legend4_name) at ($(legend3_name.east) + (3em, 0)$) {\centering Low \\ Frequency};
\end{tikzpicture}

%% file: shortDist.tex
\section{Trajectory Servoing}\label{sec:shortDist}

\subsection{System Overview
\label{sec:BG_Overview}}
The algorithmic components and information flow of a trajectory
servoing system are depicted in Fig.~\ref{fig:system_flowchart}, and
consist of two major components.  The first one, described in 
\S\ref{sec:shortDist}, is a trajectory servoing system for a set of
world points and specified trajectories.  These points are obtained from
the V-SLAM system as well as tracked over time.  It is capable of
guiding a mobile robot along short paths. The second component,
described in \S\ref{sec:longDist}, supervises the core trajectory
servoing system to confirm that it has sufficient features from
the feature pool to operate.  Should this quantity dip too low, the
supervisor queries the V-SLAM module for additional features and the
robot pose to build new feature tracks.


\subsection{Trajectory Servoing}
\label{sec:BG_trajServo}
{This section describes the basic {\em trajectory servoing}
implementation. Traditional controllers are designed in Cartesian space
where V-SLAM is a pose observer (also called visual odometry). 
For short distance navigation, where sufficient image features remain
within the field-of-view (FoV) over the trajectory, we shift the
problem to image space and solve using IBVS by synthesizing a desired
feature trajectory that defines what the camera should see over time
from an initial view.}

\begin{figure}[t]
  \vspace*{0.1in}
\centering
  \begin{tikzpicture}[inner sep=0pt, outer sep=0pt, auto, >=latex]
  \node[anchor=south west] (fig) at (0,0)
    	{{\scalebox{0.65}{\input{figs/trajservo_demo.tex}}}};
  \end{tikzpicture}
  \caption{The trajectory servoing process uses matches from
  $\featureSet^*$ to $\featureSet$ to define the control $\control$, where
  $\featureSet^*(t)$ is computed from the desired trajectory $\mcframe{\roboF}{W,*}{R}(t)$.\label{fig:trajservo_demo}}

  \vspace*{-1em}
\end{figure}

The standard IBVS equations presented in \S\ref{sec:ibvs} typically
apply to tracked features with known static positions in the world
(relative to some frame attached to these positions). 
As described in \S \ref{sec:BG_vs} and \S \ref{sec:BG_vtr}, a visual map
or visible targets are usually necessary.
The prerequisites needed for stable tracking and depth recovery of
features are major challenges regarding the use of visual servoing 
in unknown environments.  Fortunately, they are all realizable based on
information and modules available within mobile robot autonomy stacks.

Trajectory servoing requirements condense down to the following: 
{\bf 1)} A set of image points, $\featureSet^*(t_0)$, with known (relative) 
positions; 
{\bf 2)} A given trajectory and control signal for the robot starting at the
robot's current pose or nearby, $\mcframe{\roboF}{W}{R}(t_0)$; and 
{\bf 3)} The ability to index and associate the image points across future
image measurements, $\featureSet^*(t) \leftrightarrow \featureSet(t)$,
when tracking the trajectory. 
In other words, it requires a mechanism to temporally associate 
measured features along the entire trajectory.
The trajectory servoing process and
variables are depicted in Fig.  \ref{fig:trajservo_demo}.  The autonomy
modules contributing this information are the navigation and V-SLAM
stacks. The navigation stack generates a trajectory to follow. 
An indirect, feature-based V-SLAM stack keeps track of points in the
local environment, links them to previously observed visual features,
and estimates their actual positions relative to the robot.

\subsubsection{Trajectory and control signals}
Define $\featureSet = \set{\camPt_i}_{1}^{\numFeat} \subset \Real^2$ as a set of image points in the current camera image, sourced from the set $\pointSet = \set{\pointIndividual_i^{\mc W}}_{1}^{\numFeat} \subset \Real^3$ 
of points in the world frame.  Suppose that the robot should attain a future pose given by $\roboF^*$, for which the points in $\pointSet$ will project to the image coordinates $\featureSetD = \camProj \circ
(\roboF^* \mcframe{\camF}{R}{C})^{-1}(\pointSet)$. For simplicity,
ignore field of view issues and occlusions between points. Their effect
would be such that only a subset of the points in $\pointSet$ would
contribute to visual servoing.

Assume that a specific short-duration path has been established as the
one to follow, and has been converted into a path relative to the
robot's local frame. It either contains the current robot pose in it, or
has a nearby pose. Contemporary navigation stacks have a means to
synthesize both a time-varying trajectory and an associated control
signal from the paths. Here, we apply a standard trajectory tracking
controller \cite{NITraj} to generate $\roboBV^*(t)$ and
$\mcframe{\roboF}{W,*}{R}(t)$ by forward simulating \eqref{eq:roboDE};
note that $\roboBV^*$ contains the linear velocity $\nu^*$ and angular
velocity $\omega^*$. Some navigation stacks use optimal control
synthesis to build the trajectory. 
Either way, the generated trajectory is achievable by the robot.

In the time-varying trajectory tracking case, we assume that a trajectory reference $\mcframe{\camF}{W}{C}(t)$ exists along with a control signal $\roboVelDes (t)$ satisfying \eqref{eq:camDE}. It would typically be derived from
a robot trajectory reference $\mcframe{\roboF}{W}{R}(t)$ and control signal $\roboVelDes (t)$ satisfying \eqref{eq:roboDE}. Using those time-varying functions, the equations in \eqref{eq:camDE} are solved to obtain the image coordinate trajectories. Written in short-hand to expose only the main variables, the forward integrated feature trajectory $\featureSetD$ is:
\begin{equation}  \label{eq:featDE}
\begin{split}
  \dot \featureSetD & = \imJac \circ \mcframe{h}{C}{W}(t)(\pointSet^{\mc W}) \cdot
    \zeta_{\roboVelDes (t)}, \ \text{with} \\
    \featureSetD(0) & = \camProj \circ \mcframe{h}{C}{W}(0)(\pointSet^{\mc W}).
\end{split}
\end{equation}
It will lead to a realizable visual servoing problem where $\nu^*$, $\omega^*$, and $\featureSetD(t)$ are consistent with each other. The equations will require converting the reference robot trajectory to a camera trajectory $\mcframe{\camF}{W,*}{C}(t)$ using \raisebox{1pt}{$\Ad^{-1}_{\mcframe{h}{R}{C}}$}.
\vspace{4pt}
\subsubsection{Features and feature paths\label{FeatPaths}}
The V-SLAM module provides a pool of visible features with known relative position for the current stereo frame, plus a means to assess future visibility if desired. Taking this pool to define the feature set $\featureSetD(0)$ gives the final piece of information
needed to forward integrate \eqref{eq:featDE} and generate feature
trajectories $\featureSetD(t)$ in the left camera frame. This process
acts like a short-term teach and repeat feature trajectory planner but
is {\em simulate} and repeat, for on-the-fly generation of the repeat data. 

A less involved module could be used besides a fully realized V-SLAM system, however doing so would require creating many of the fundamental building blocks of an indirect, feature-based V-SLAM system. 
Given the availability of strong performing open-source, real-time V-SLAM methods, there is little need to create a custom module. 
In addition, an extra benefit to tracked features through V-SLAM system
is that a feature map is maintained to retrieve same reappeared
features. As will be shown, this significantly improves the average
lifetime of features, especially compared to a simple frame by frame
tracking system without the feature map. 

After the V-SLAM feature tracking process, we are already working with this feasible set whereby the indexed elements in $\featureSet$ correspond exactly to their counterpart in $\featureSetD$ with the same index, i.e., the sets are {\em in correspondence}.
\subsubsection{Trajectory Servoing Control}
Define the error to be $\errorSet = \featureSet - \featureSetD$ where
elements with matched indices are subtracted. The error dynamics of the
points are:
\begin{equation} \label{eq:ErrorSetDE}
  \dot \errorSet = \dot \featureSet - \dot \featureSetD = 
    \imJac_{\roboVel}(\mcframe{\camF}{C}{W}(\pointSet), \featureSet; \mcframe{h}{R}{C}) \cdot \roboVel - \dot \featureSetD,
\end{equation}
where we apply the same argument adjustment as in \eqref{eq:imJac} so
that dependence is on image features then point coordinates as needed.
Further, functions or operations applied to indexed sets will return an
indexed set whose elements correspond to the input elements from the
input indexed set. Since the desired image coordinates $\featureSetD$
are not with respect to a static goal pose but a dynamic feature
trajectory, $\dot \featureSetD \neq 0$, see \eqref{eq:featDE}.
Define $\error$, $\featSetV$, $\featSetDV$, $\pointSetV$ 
and $\imJacV$ to be the
vectorized versions
of $\errorSet$, $\featureSet$, $\featureSetD$, $\pointSet$ and $\imJac$. Then,
\begin{equation} \label{eq:errorDE}
  \dot \error =
  \imJacV(\mcframe{\camF}{C}{W}(\pointSetV),\featSetV;\mcframe{h}{R}{C}) \cdot \roboVel
    - \featSetDVdot
\end{equation}
is an overdetermined set of equations for $\control$ when $\numFeat \geq 2$.
$\imJacV = [\imJacV^1, \imJacV^2] \in \Real^{2\numFeat \times 2}$.
Removing the functional dependence and breaking apart the different
control contributions, the objective is to satisfy,
\begin{equation} \label{eq:IBVSw}
  \dot \error = \imJacV \cdot \roboVel-\featSetDVdot = \imJacV^1 \nu + \imJacV^2 \omega -\featSetDVdot
              = -\lambda \error.
\end{equation}
A least-squares solution establishes the angular rate feedback,
\begin{equation} \label{eq:vStEq}
  \omega = {\of{\imJacV^2}}^\dagger \of{ -\imJacV^1 \nu -\lambda \error +\featSetDVdot},
\end{equation}
so that
\begin{equation}
  \dot \error = -\lambda \error + \Delta \error,
\end{equation}
where $\Delta \error$ is mismatch between the true solution and the
computed pseudo-inverse solution. 
If the overdetermined linear system equalities \eqref{eq:IBVSw} are
compatible and have a unique solution \eqref{eq:vStEq}, then
$\Delta \error$ will vanish and the robot will achieve the target pose.
If $\Delta \error$ does not vanish, then there will be an error (usually
some fixed point $\error_{ss} \neq 0$). 
For mobile robots, it is common to use the linear velocities from
$\nu^*(t)$ of the given trajectory \cite{SEGVIC2009172,inproceedingsNwL}
for angular control \eqref{eq:vStEq}. 
The decoupling provides robustness to the motion imperceptibility 
problem that can affect translational motion control \cite{4141039, sharma1997motion}. 

The vectorized form of \eqref{eq:featDE} for $\featureSetD(t)$ is 
\begin{eqnarray}
\featSetDVdot=\imJacV^1(\pointSetV^{\mc C^*}(t),\featSetDV(t)) \nu^* {(t)} +\imJacV^2(\pointSetV^{\mc C^*}(t),\featSetDV(t)) \omega^* {(t)}.
\end{eqnarray} 
Continuing, the vectorized steering equation \eqref{eq:vStEq} leads to
\begin{multline} \label{eq:IBVStrajTrack}
  \omega = {\of{\imJacV^2(\pointSetV^{\mc C},\featSetV)}}^\dagger
    \bigg( \imJacV^1(\pointSetV^{\mc C^*}(t),\featSetDV(t)) \nu^* {(t)} 
          -\imJacV^1(\pointSetV^{\mc C},\featSetV) \nu
    \\
      + \imJacV^2(\pointSetV^{\mc C^*}(t),\featSetDV(t)) \omega^* {(t)}
      - \lambda \error \bigg).
\end{multline}
They consist of feedforward terms derived from the desired trajectory and
feedback terms to drive the error to zero. The feedforward terms should cancel
out the $\dot \featureSetD$ term in \eqref{eq:ErrorSetDE}, or
equivalently the now non-vanishing $\featSetDVdot$ term in 
\eqref{eq:errorDE}. 
When traveling along the feature trajectory $\featureSetD(t)$, the
angular velocity $\omega$ is computed from \eqref{eq:IBVStrajTrack},
where starred terms are known, and $\nu=\nu^*(t)$ from discussion
after \eqref{eq:vStEq}.
As far as we know, no general IBVS tracking equations have been derived that
combine feedforward and feedback signals.

%% file: figs/trajservo_demo.tex
\tikzstyle{newline} = [-, thick]
\tikzstyle{newline_dashed} = [-, thick, dashed]
\tikzstyle{newline_dotted} = [-, thick, dash pattern=on 1pt off 1pt on 1pt]
\tikzstyle{newtip} = [->, thick]
\tikzstyle{newtip_dashed} = [->, thick, dashed]
\tikzstyle{block} = [draw, rectangle, dashed, thick,rounded corners=2pt,
                     minimum height=1.5em, minimum width=5em, inner sep=4pt]

\newcommand{\figPose}[1]
{
\begin{tikzpicture}
  \node (r1) at (0,0) {};
  \node (r2) at (3,-1) {};
  \node (r3) at (0,-2) {};
  
  \draw[very thick] (r1.center) -- (r2.center) -- (r3.center) -- cycle;
  
  \fill[fill=#1] (r1.center)--(r2.center)--(r3.center);
\end{tikzpicture}
}

\tikzstyle{features}=[star, draw, inner sep=1pt]

\begin{tikzpicture}[inner sep=0pt, outer sep=0pt, auto, >=latex]
  \node[anchor=west, scale=0.3] (refp1) at (0,0) {\figPose{red!60}};
  \node[anchor=west, scale=0.3] (refp2) at ($(refp1.east) + (0.7in, 0in)$) {\figPose{red!60}};
  \node[anchor=west, scale=0.3] (refp3) at ($(refp2.east) + (0.7in, 0in)$) {\figPose{red!60}};
  
  \node[anchor=north, scale=0.3] (realp1) at ($(refp1.south) + (0in, -1in)$) {\figPose{gray!40}};
  \node[anchor=north, scale=0.3] (realp2) at (realp1.north-|refp2.south) {\figPose{gray!40}};
  \node[anchor=north, scale=0.3] (realp3) at (realp1.north-|refp3.south) {\figPose{gray!40}};
  
  \node[features, fill=red!60, anchor=north, scale=2] (reff1) at ($(refp1.south) + (0in, -0.4in)$) {};
  \node[features, fill=red!20, anchor=north, scale=2] (reff2) at (reff1.north-|refp2.south) {};
  \node[features, fill=red!20, anchor=north, scale=2] (reff3) at (reff1.north-|refp3.south) {};
  
  \node[features, fill=gray!40, anchor=south, scale=2] (curf1) at ($(realp1.north) + (0in, 0.3in)$) {};
  \node[features, fill=gray!40, anchor=south, scale=2] (curf2) at (curf1.south-|realp2.north) {};
  \node[features, fill=gray!40, anchor=south, scale=2] (curf3) at (curf1.south-|realp3.north) {};
  
  \draw[newline] ($(refp1.east) + (0.05in,0in)$) -- ($(refp1.east) + (0.2in,0in)$);
  \draw[newline_dotted] ($(refp1.east) + (0.2in,0in)$) -- ($(refp2.west) + (-0.3in,0in)$);
  \draw[newtip] ($(refp2.west) + (-0.3in,0in)$) -- ($(refp2.west) + (-0.1in,0in)$);
  \draw[newtip] ($(refp2.east) + (0.05in,0in)$) -- ($(refp3.west) + (-0.1in,0in)$);
  
  \draw[newline] ($(refp1.south) + (0in,0in)$) -- ($(reff1.north) + (0in,0.1in)$);
  \draw[newtip_dashed] ($(reff1.north) + (0in,0.1in)$) -- ($(refp2.south) + (0in,0in)$) {};
  \draw[newtip_dashed] ($(reff1.north) + (0in,0.1in)$) -- ($(refp3.south) + (0in,0in)$) {};
  
  \draw[newtip_dashed] ($(refp2.south) + (0in,0in)$) -- ($(reff2.north) + (0in,0.1in)$);
  \draw[newtip_dashed] ($(refp3.south) + (0in,0in)$) -- ($(reff3.north) + (0in,0.1in)$);
  
  \draw[newline] ($(realp1.north) + (0in,0in)$) -- ($(curf1.south) + (0in,-0.1in)$);
  \draw[newline] ($(realp2.north) + (0in,0in)$) -- ($(curf2.south) + (0in,-0.1in)$);
  \draw[newline] ($(realp3.north) + (0in,0in)$) -- ($(curf3.south) + (0in,-0.1in)$);
  
  \node[block, anchor=north, minimum width=0.18in, minimum height=0.4in] (box1) at ($(reff1.north) + (0in,0.05in)$) {}; 
  \node[block, anchor=north, minimum width=0.18in, minimum height=0.4in] (box2) at ($(reff2.north) + (0in,0.05in)$) {}; 
  \node[block, anchor=north, minimum width=0.18in, minimum height=0.4in] (box3) at ($(reff3.north) + (0in,0.05in)$) {}; 
  
  \draw[newline] (box1.east) -- ($(box1.east) + (0.25in,-0.195in)$);
  \draw[newline_dotted] ($(box1.east) + (0.25in,-0.195in)$) -- ($(realp2.west) + (-0.3in,0.195in)$);
  \draw[newtip] ($(realp2.west) + (-0.3in,0.195in)$) -- ($(realp2.west) + (-0.05in,0in)$) node[midway,right,xshift=-12pt,yshift=13pt]{$\control(t_i)$} ;
  \draw[newtip] (box2.east) -- ($(realp3.west) + (-0.05in,0in)$) node[midway,right,xshift=5pt]{$\control(t_{i+1})$};
  
  \node[anchor=south] (time1) at ($(refp1.north) + (0in,0in)$) {$t_0$};
  \node[anchor=south] (time2) at ($(refp2.north) + (0in,0in)$) {$t_i$};
  \node[anchor=south] (time3) at ($(refp3.north) + (0in,0in)$) {$t_{i+1}$};
  
  \node[anchor=west] (reftext) at ($(refp3.east) + (0.1in,0in)$) {$\mcframe{\roboF}{W,*}{R}(t)$};
  \node[anchor=north] (refftext) at (reff3.north-|reftext.south) {$\featureSetD(t)$};
  \node[anchor=north] (curftext) at (curf3.north-|refftext.south) {$\featureSet(t)$};
  \node[anchor=north,yshift=-0.5em] (realptext) at (realp3.north-|curftext.south) {\centering $\mcframe{\roboF}{W}{R}(t)$};
  
  \node[anchor=north] (proj1) at ($(refp1.east) + (0.35in,-0.45in)$) {\footnotesize $\camProj \circ \mcframe{h}{C}{W}(t_i)$};
  \node[anchor=north] (proj2) at ($(refp2.east) + (0.35in,-0.45in)$) {\footnotesize $\camProj \circ \mcframe{h}{C}{W}(t_{i+1})$};
  
  \node[anchor=east] (q) at ($(reff1.west) + (0in,0.2in)$) {$\pointSet^{\mc W}$};
  
  \node[anchor=east, text centered, text width=5em] (desired) 
  at ($(refp1.west) + (-1.2em, -2em)$) 
  {\centering \large \sc Desired};
  \node[anchor=east, text centered, text width=5em] (actual) 
  at ($(realp1.west) + (-1.2em, 1.2em)$) 
  {\centering \large \sc Actual};
  
  \draw[-,very thick] ($(box1.west)+(-7em,-0.15em)$) -- ($(box1.west)+(-1.8em,-0.15em)$);
  
  \node[anchor=west, text centered, text width=5em] (dp) 
  at ($(reftext.east) + (0em, 0em)$) 
  {\centering \sc Pose};
  \node[anchor=center, text centered, text width=5em,yshift=0.5em] (df) 
  at (dp.south|-refftext.east) 
  {\centering \sc Feature};
  \node[anchor=center, text centered, text width=5em,yshift=-0.5em] (af) 
  at (df.south|-curftext.east)
  {\centering \sc Feature};
  \node[anchor=center, text centered, text width=5em] (ap) 
  at (af.south|-realptext.east) 
  {\centering \sc Pose};
  
  \draw[-,very thick] ($(box3.east)+(9.3em,-0.15em)$) -- ($(box3.east)+(4.9em,-0.15em)$);

\end{tikzpicture}

%% file: shortExp.tex
\begin{figure*}[t]
  \begin{minipage}[t]{0.3\textwidth}
  \centering
  {\includegraphics[height=0.9in]{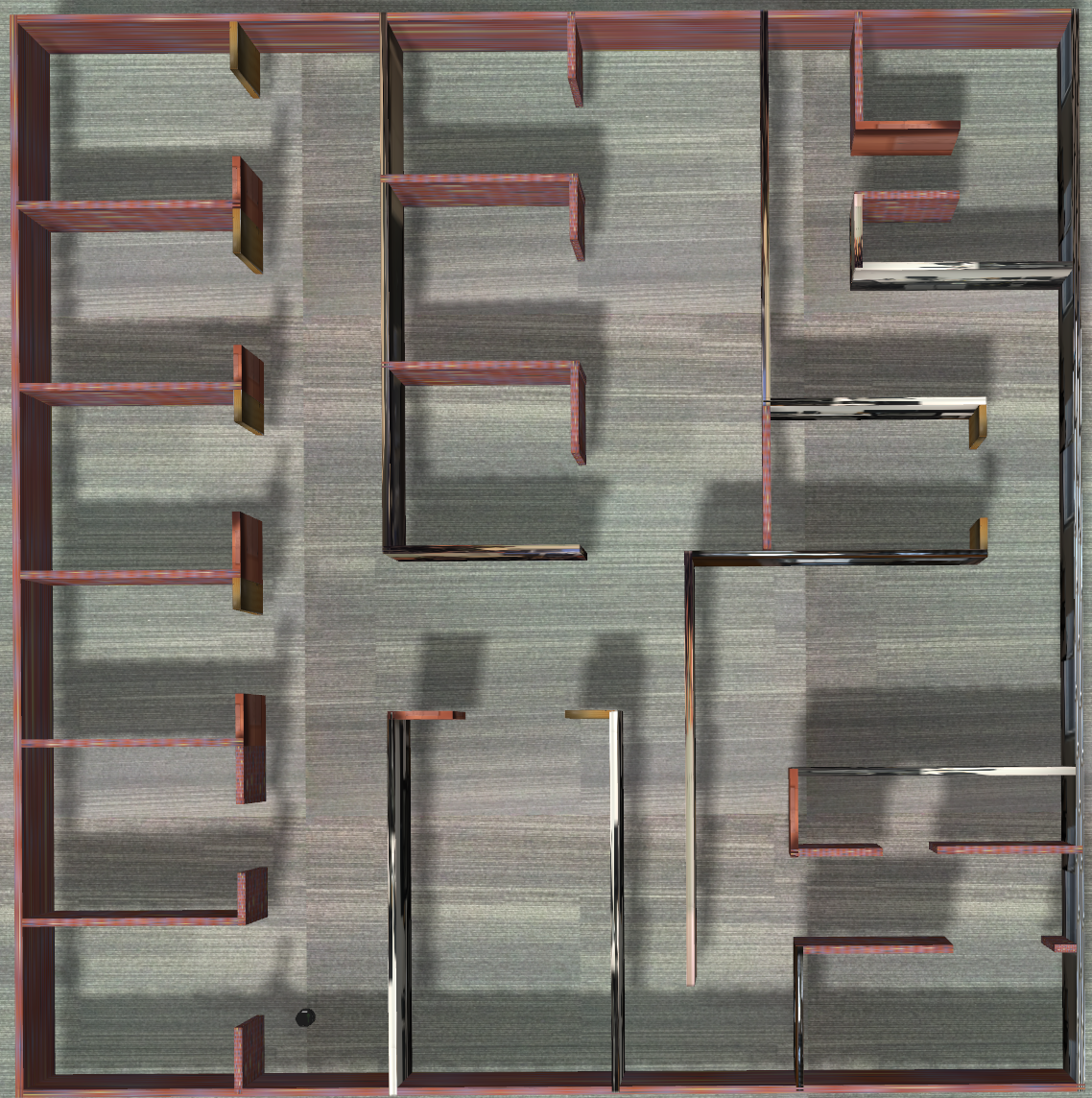}}
  {\includegraphics[height=0.9in]{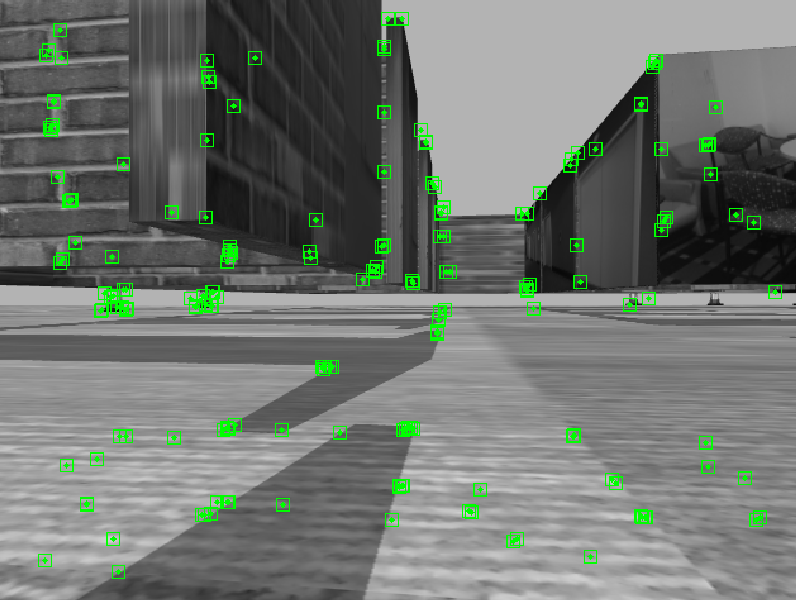}}
  \caption{Gazebo environment top view and robot view with SLAM features.\label{fig:tsrb_env}}
  \end{minipage}
  \hfill
  \begin{minipage}[t]{0.2\textwidth}
  \centering
  \begin{tikzpicture}[inner sep=0pt, outer sep=0pt]
    \node (S) at (0in,0in) 
      {\includegraphics[scale=0.06]{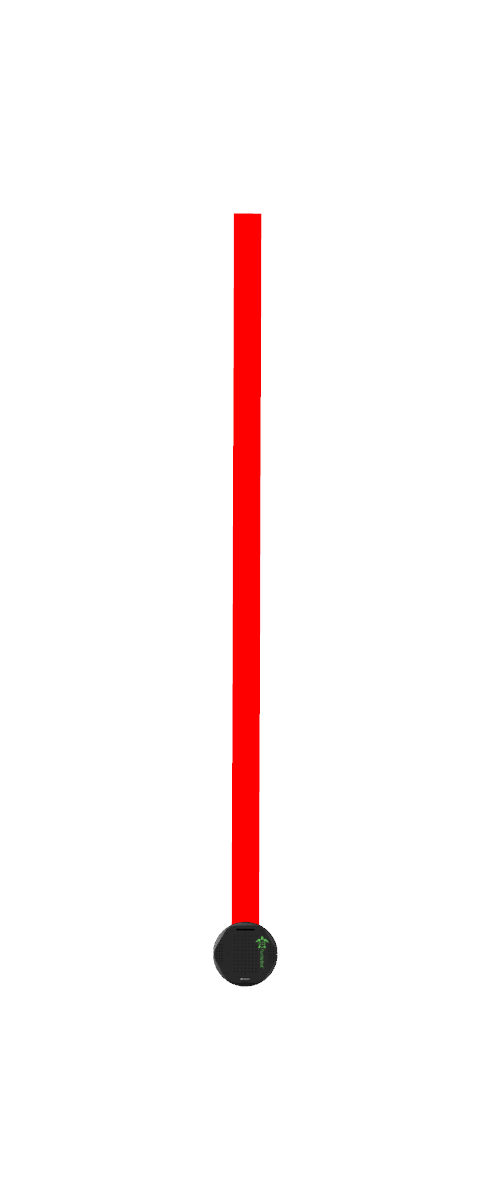}};
    \node[anchor=south west,xshift=-10pt] (WT) at (S.south east)
      {\includegraphics[scale=0.06]{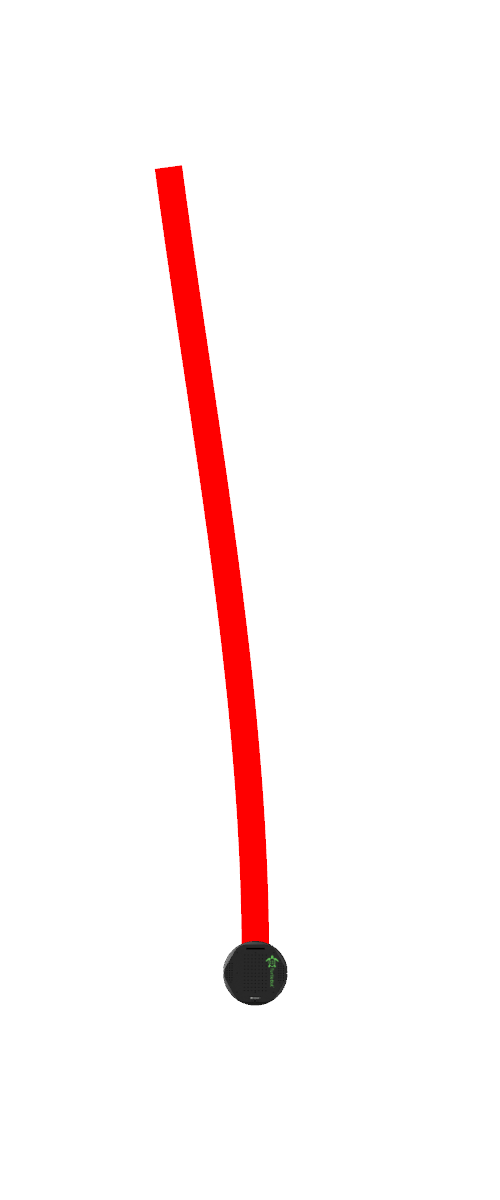}};
    \node[anchor=south west,xshift=-10pt] (ST) at (WT.south east)
      {\includegraphics[scale=0.06,clip=true,trim=0in 0in 0in 0in]{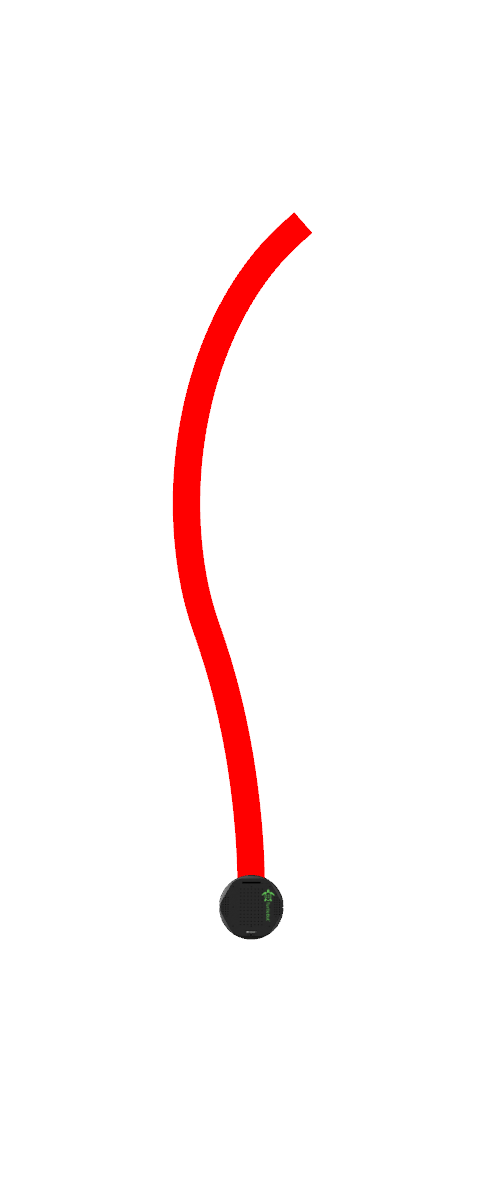}};
    \node[anchor=south west,,xshift=-10pt] (TS) at (ST.south east)
      {\includegraphics[scale=0.06,clip=true,trim=0in 0in 0in 0in]{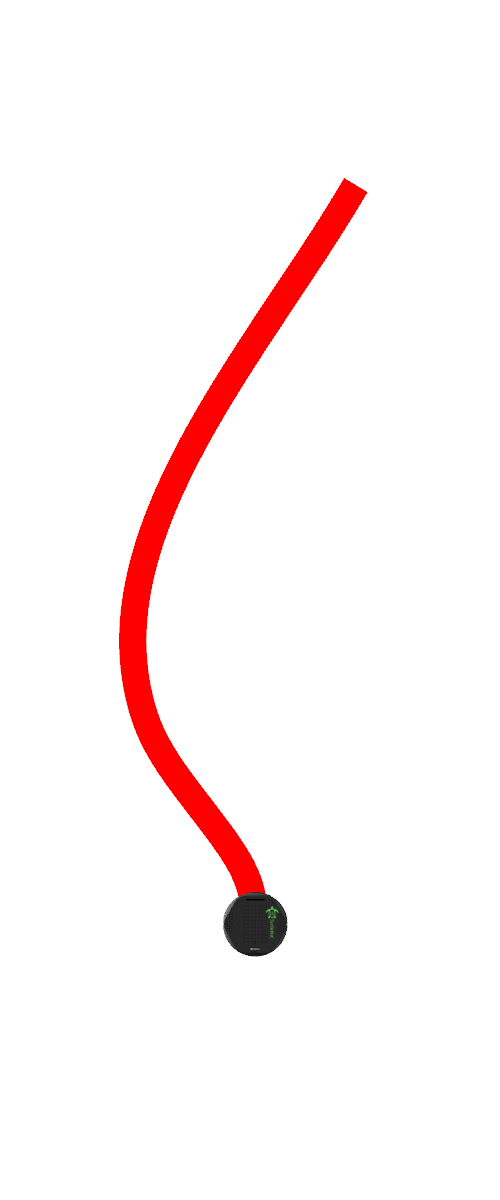}};
    \node[anchor=south west,,xshift=-8pt] (TT) at ($(TS.south east) + (0, 0.15in)$)
      {\includegraphics[scale=0.042,clip=true,trim=0in 0in 0in 0in]{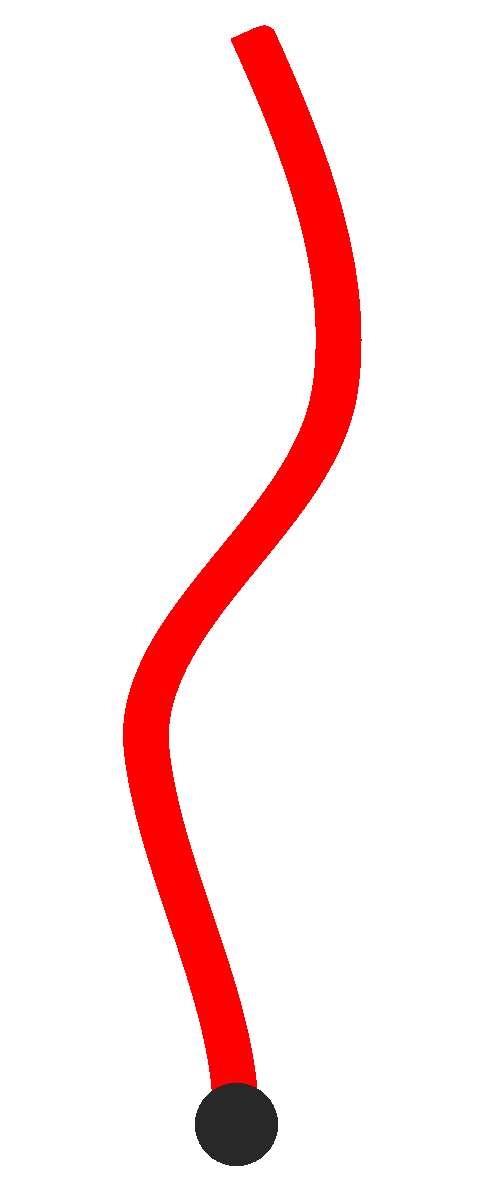}};

  \node[anchor=north] at ($(S.south)+(0in,6pt)$) {\footnotesize SS};
  \node[anchor=north] at ($(WT.south)+(0in,6pt)$) {\footnotesize SWT};
  \node[anchor=north] at ($(ST.south)+(0in,6pt)$) {\footnotesize SST};
  \node[anchor=north] at ($(TS.south)+(0in,6pt)$) {\footnotesize STS};
  \node[anchor=north] at ($(TT.south)+(0in,-5pt)$) {\footnotesize STT};
  \end{tikzpicture}
  \caption{Short distance template trajectories.  \label{fig:short_paths}}
  \end{minipage}
  \hfill
  \begin{minipage}[t]{0.45\textwidth}
  \centering
  {\includegraphics[height=0.9in]{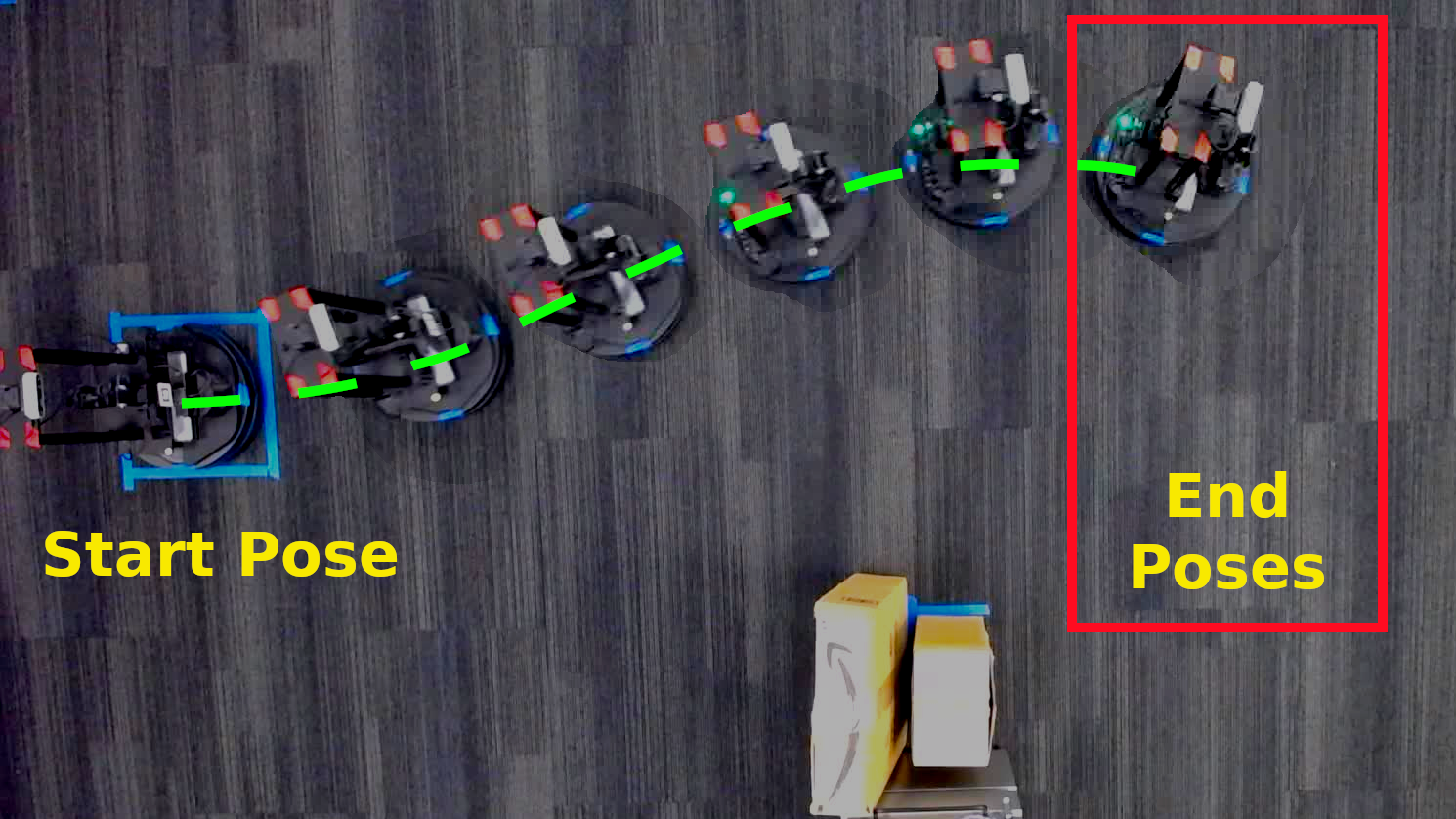}}
  {\includegraphics[height=0.9in]{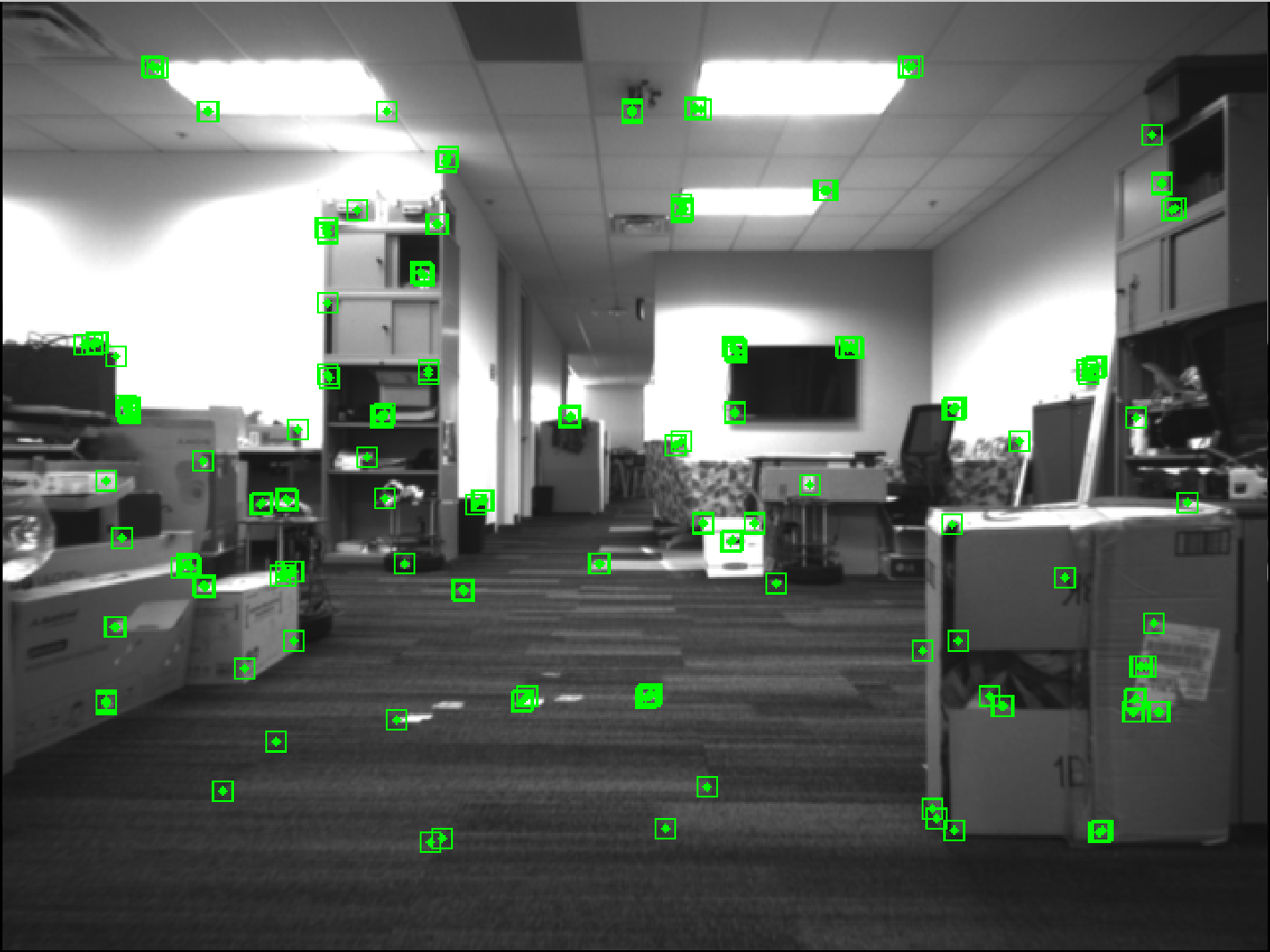}}
  \caption{Real experiment top view and robot view with SLAM features. Blue box is the robot's start pose. Red box shows the end poses region of short trajectories. The green curve is a sample trajectory to track.\label{fig:real_env}}
\end{minipage}
  \vspace*{-1em}
\end{figure*}

\subsection{Simulation Experiments and Results}\label{sec:shortDist_exp}

This section runs several short distance trajectory servoing
experiments to evaluate the accuracy of the image-based feedback
strategy supplemented by stereo V-SLAM. The hypothesis is that
short distance trajectory tracking in image space will improve over
tracking in pose space.

\subsubsection{Experimental Setup}
For quantifiable and reproducible outcomes, the ROS/Gazebo SLAM
evaluation environment from \cite{zhao2020closednav} is used for
the tests, on an Intel i7-8700 workstation. 
Fig. \ref{fig:tsrb_env} shows a top down view of the world
plus a robot view. The simulated robot is a Turtlebot.  
It is tasked to follow a given short distance trajectory, whose desired
linear velocity is $0.3m/s$ and which is generated to be dynamically
feasible \cite{NITraj}. 
A total of five paths were designed, loosely based on Dubins paths. They
are denoted as short: straight (SS), weak turn (SWT), straight+turn (SST),
turn+straight (STS), turn+turn (STT), and are depicted in Fig.
\ref{fig:short_paths}. The average trajectory lengths are $\sim$4m.
Longer paths would consist of multiple short segment reflecting
variations on this path set.  They are designed to ensure that
sufficient feature points, visible in the first frame, remain visible
along the entirety of the path. 
Five trials per trajectory are run.  The
desired and actual robot poses are recorded for performance scoring.

Two metrics quantify tracking performance:
a) Average lateral error (ALE) is the 2-norm of the perpendicular
distance to the robot heading direction averaged over time.
It measures robot deviation from the desired trajectory and has been
used to evaluate steering controllers \cite{dominguez2016comparison};
b) Terminal error (TE) measures the robot's distance to the final stopped
position of the desired trajectory after tracking ends.  Implementation
details are provided in the public Github repository \cite{gitTS}.

\subsubsection{SLAM Stack}
Part of the robot's software stack includes the Good Feature (GF) ORB-SLAM
system \cite{8964284} for estimating camera poses. It is configured to
work with a stereo camera and integrated into a loosely coupled,
visual-inertial (VI) system \cite{zhao2020closednav,20.500.11850/52698} based on a 
multi-rate filter to form a VI-SLAM system. The trajectory servoing system will interface with
the GF-ORB-SLAM system to have access to tracked features for servoing.

\makeatletter
\newcommand*\bigcdot{\mathpalette\bigcdot@{.7}}
\newcommand*\bigcdot@[2]{\mathbin{\vcenter{\hbox{\scalebox{#2}{$\m@th#1\bullet$}}}}}
\makeatother

\begin{figure*}
\begin{minipage}[t]{0.55\textwidth}
\vspace*{-1.45in}
\captionof{table}{ Short distance trajectory benchmark and real experiment results.\label{tab:shortTrajResults}}
\hspace*{-0.15in}
\begin{tikzpicture}[inner sep=0pt,outer sep=0pt,scale=1, every node/.style={scale=0.8}]

	\node[anchor=north west] (sim_ale) at (0, 0pt)
    {
    \setlength{\tabcolsep}{2pt}
    \begin{tabular}{|c||c|ccc|}
    \hline 
    \textbf{Seq.} & PO & SLAM & TS & VS+ \\ 
    \hline 
    SS  & 0.70  & 0.84  & 0.88  & x \\ 
    SWT & 0.71  & 1.35  & 1.23  & x \\ 
    SST & 0.55  & 1.64  & 0.82  & x \\ 
    STS & 0.86  & 1.68  & 0.91  & x \\ 
    STT & 0.75  & 1.21  & 0.96  & x \\ 
    \hline 
    \textbf{Avg.} & 0.71 & 1.34 & \textbf{0.96} & x \\ 
    \hline 
    \end{tabular}
    };
    
    \node[anchor=south, text width=5cm, text centered] (sim_ale_cap) 
    at ($(sim_ale.north) + (0pt, 2pt)$)
    {\normalsize \textbf{(a)} Sim {ALE} (cm)};
      
    \node[anchor=west] (sim_te) at ($(sim_ale.east) + (2pt, 0)$)
    {
    \setlength{\tabcolsep}{4pt}
    \begin{tabular}{|c||c|cc|}
    \hline 
    \textbf{Seq.} & PO & SLAM & TS \\ 
    \hline 
    SS  & 1.06  & 1.53  & 2.05  \\ 
    SWT & 1.24  & 2.23  & 2.87  \\ 
    SST & 2.07  & 3.33  & 2.47  \\ 
    STS & 1.97  & 4.14  & 2.18  \\ 
    STT & 4.34  & 4.99  & 2.42  \\ 
    \hline 
    \textbf{Avg.} & 2.14 & 3.24 & \textbf{2.40} \\ 
    \hline 
    \end{tabular}
    };
    
	\node[anchor=south, text centered] (sim_te_cap) 
    at ($(sim_te.north) + (0pt, 2pt)$)
    {\normalsize \textbf{(b)} Sim Terminal Error (cm)};    
    
    \node[anchor=west] (real_te) at ($(sim_te.east) + (2pt, 0pt)$)
    {
    \setlength{\tabcolsep}{4pt}
    \begin{tabular}{|c||ccc|}
    \hline 
    \textbf{Seq.} & RO & SLAM & TS \\ 
    \hline 
    SS  & 4.5  & 8.9  & 4.9  \\ 
    SWT & 6.8  & 10.4  & 4.9  \\ 
    SST & 8.0  & 13.1  & 5.7  \\ 
    STS & 13.3  & 11.9  & 5.0  \\ 
    STT & 9.5  & 10.5  & 3.5  \\ 
    \hline 
    \textbf{Avg.} & 8.4 & 10.9 & \textbf{4.8} \\ 
    \hline 
    \end{tabular}
    };
	
	\node[anchor=south, text centered] (real_te_cap) 
    at ($(real_te.north) + (0pt, 2pt)$)
    {\normalsize \textbf{(c)} Real Terminal Error (cm)};      

\end{tikzpicture}
  
\end{minipage}
\vspace*{-1em}
\hspace*{1em}
\begin{minipage}[b]{0.4\textwidth}
\vspace*{0.01in}
\centering
  \begin{tikzpicture}[inner sep=0pt, outer sep=0pt, auto, >=latex]
%
  
  \node[anchor=south west] (ale) at (0,0)
    	{{\scalebox{0.3}{\includegraphics[width=1.7in,clip=true,trim=0in 0in 0in 0in]{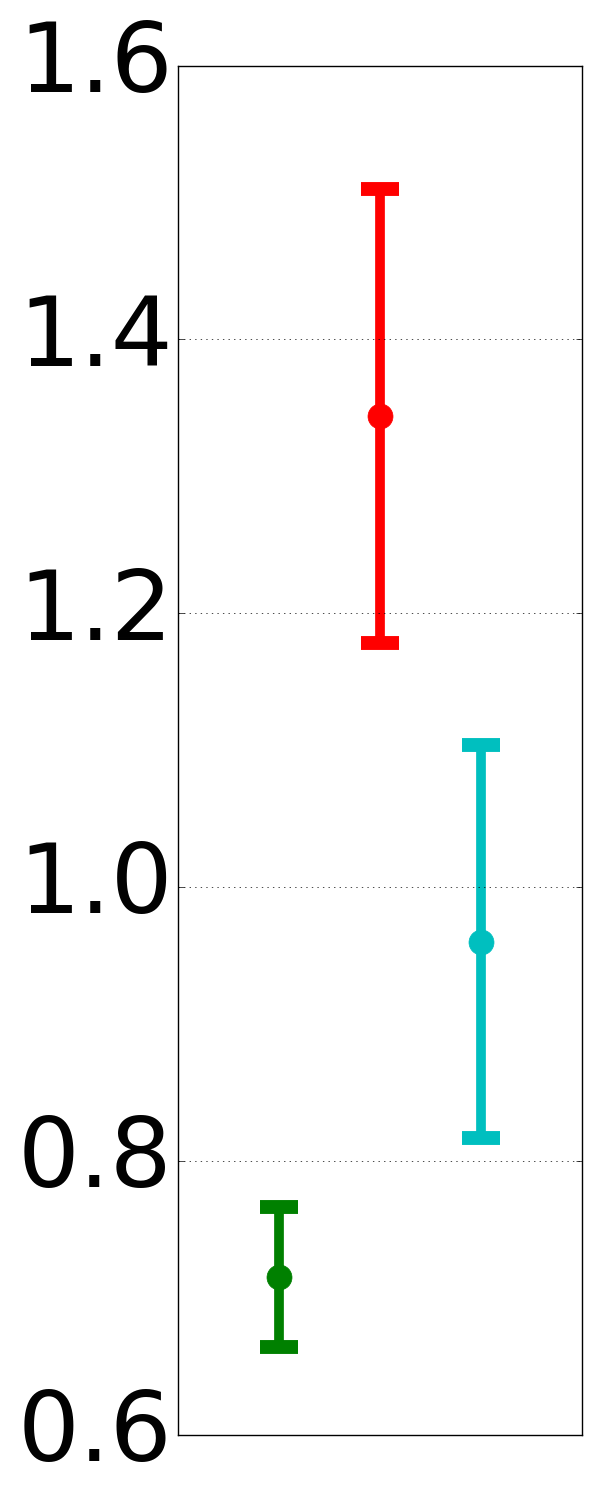}}}};
	
  \node[anchor=south, rotate=90] (col_yaxis_ale) at ($(ale.west) + (-2pt, 0)$) {\footnotesize Sim ALE (cm)};

  \node[anchor=west] (te) at ($(ale.east) + (12pt, 0)$)
    	{{\scalebox{0.3}{\includegraphics[width=1.7in,clip=true,trim=0in 0in 0in 0in]{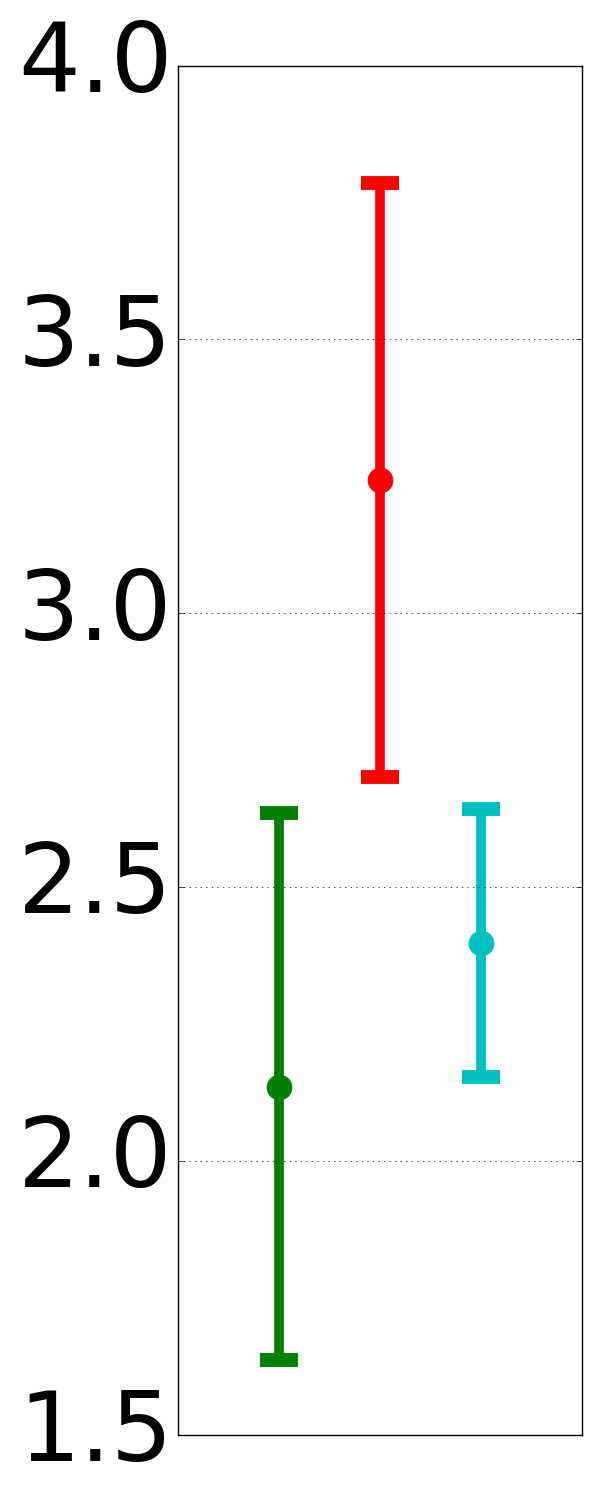}}}};
	
  \node[anchor=south, rotate=90] (col_yaxis_te) at ($(te.west) + (-2pt, 0)$) {\footnotesize Sim Terminal Error (cm)};
  
  \node[anchor=west] (rte) at ($(te.east) + (12pt, 0)$)
    	{{\scalebox{0.3}{\includegraphics[width=1.7in,clip=true,trim=0in 0in 0in 0in]{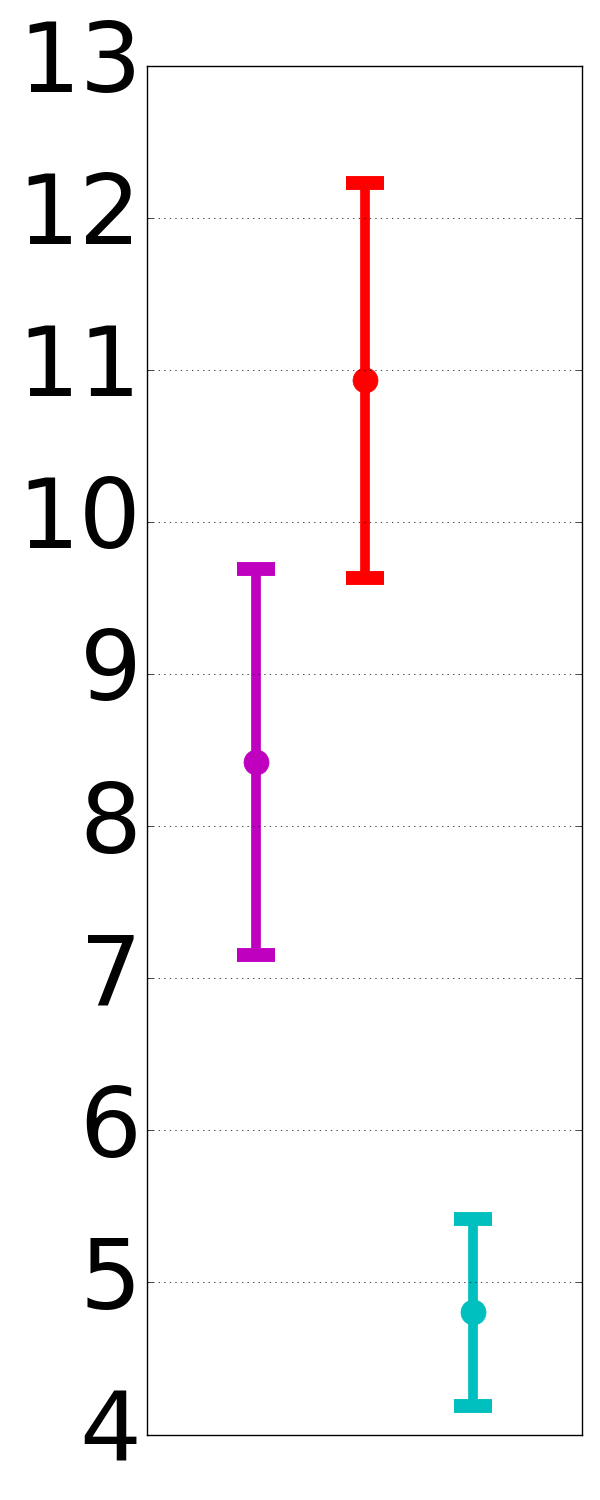}}}};
	
  \node[anchor=south, rotate=90] (col_yaxis_rte) at ($(rte.west) + (-2pt, 0)$) {\footnotesize Real Terminal Error (cm)};
  
  \node[anchor=west] (legend) at ($(rte.east) + (-1pt, 28pt)$)
    	{{\scalebox{0.25}{\includegraphics[width=1.7in,clip=true,trim=0in 0in 0in 0in]{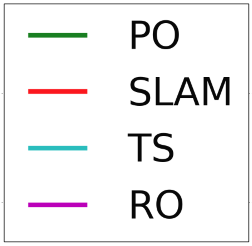}}}};
  \end{tikzpicture}
  \vspace*{-0.5em}
  \caption{95\% confidence intervals for short distance outcomes. 
  \label{fig:shortTrajBench}}
\end{minipage}
\end{figure*}

\subsubsection{Methods Tested}
The baseline performance standard is pose-based control using {\em
perfect odometry} (PO) as obtained from the actual robot pose in the
Gazebo simulator. PO is used for performance comparison of the tested
methods.  Two comparison methods are implemented.  The first replaces PO
with the V-SLAM estimated pose (SLAM).  The second is an implementation of
IBVS based on \eqref{eq:IBVStrajTrack}, which is effectively trajectory
servoing without the V-SLAM system. 
It is called VS+ to differentiate from trajectory servoing (TS), and uses a 
frame-by-frame stereo feature tracking system \cite{stvo_repo}. 
Pose-based trajectory tracking \cite{zhao2020closednav} uses a geometric
controller with feedforward $[\nu^{*},\omega^{*}]^T$ and feedback
signals:
\begin{equation} \label{eq:poseController}
\begin{split}
\nu_{cmd} & = \nu^{*}, \\
w_{cmd} &= k_{\theta}*\widetilde{\theta} + k_{y}*\widetilde{y} + \omega^{*}. \\
\end{split}
\end{equation}
where feedback uses only $\widetilde{y}$ and 
$\raisebox{-1.5pt}{$\widetilde{\theta}$}$ from the pose error,
\begin{equation} \label{eq:GErr}
  [\widetilde{x},\widetilde{y},
  \raisebox{-1.5pt}{$\widetilde{\theta}$}\makebox[2\width]{$]$}^T \simeq
  \widetilde{g} = \inverse{\roboF} \roboF^* = 
    \inverse{\of{\mcframe{\roboF}{W}{R}}}(t) 
    \of{\mcframe{\roboF}{{W},*}{R}}(t).
\end{equation}
For experimental consistency with trajectory servoing, the
pose-based control directly uses feedforward linear velocity terms 
from the given trajectory and only regulates heading.
The controller gains were empirically tuned to give good performance for
the PO case and extensively used in prior work
\cite{zhao2020closednav,Smith2017,Smith2020}.
The TS gains were also tuned \cite{gitTS}.

\subsubsection{Results and Analysis}
Tables \ref{tab:shortTrajResults}(a,b) quantify the
outcomes of all methods tested.  Fig. \ref{fig:shortTrajBench} consists
of {95\% confidence intervals of ALE and TE for the different template
trajectories and methods (minus VS+) in simulations and real experiments.}
The first outcome to note is that VS+ fails for all paths.  The average
length of successful servoing is 0.4m ($\sim$10\% of the path length).
Inconsistent data association rapidly degrades the feature pool and
prevents consistent use of features for servoing feedback.  Without the
feature map in V-SLAM, redetected features are treated as new
and assigned unique indices, which violates the {\em correspondence}
rule from \S\ref{FeatPaths};  as noted, any effort to improve this
would increasingly approach the computations found in V-SLAM.
Maintaining stable feature tracking through V-SLAM is critical to
trajectory servoing.

Using PO as the standard, the tables show a smaller gap for
TS than for SLAM as seen by the lower metric scores. For ALE, TS
experiences a 35.2\% degradation versus PO, while SLAM experiences 
an 88\% degradation.
The ALE statistics in Fig. \ref{fig:shortTrajBench} indicate that TS is
expected to outperform SLAM. Comparing PO to TS and to SLAM, the
p-values are 8e-4 and 1e-5. For TS to SLAM they are 1.9e-4. All indicate
statistically significant performance differences.  For TE, similar
results hold except that there is overlap in the PO and TS confidence
intervals. Consequently the p-value comparing PO and TS is 0.18, which
is not significant.  Thus TS performs close to PO with regards to
achieved terminal error, while SLAM does not (p-value: 2e-3).

Trajectory servoing uses the same control effort as pose-based control;
the angular control efforts of the methods were not significantly
different \cite{gitTS}. However, the rate of change of the control does
differ.  The time differentiated control signal norm is an indicator of
control smoothness (larger means less smooth). 
They are 0.174, 0.174 and 1.300 for PO, SLAM and TS respectively.
TS control is computed directly from the tracked features without
temporal regularization applied to the applied controls. 
The SLAM estimation process smooths the orientation estimates, which
translates to the control signal.

The first overall finding here is that TS outperforms SLAM and visual
servoing (VS+), though it is based on both. This confirms that TS combines
the advantages of them to enhance performance over both.
Secondly, implementing a purely image-based approach to trajectory tracking
through unknown environments is not only possible, but can work better than
SLAM pose-based controls over short segments, in the absence of global
positioning information.  The results validate the system design proposed at
the beginning of \S\ref{sec:BG_trajServo}. The robust feature tracking of V-SLAM
prevents the loss of trajectory tracking stability seen in VS+.  The V-SLAM
feature map maintenance, feature culling, and feature retrieval modules
contribute to robust visual servoing.

\subsection{Real Experiments and Results}\label{sec:shortDist_real_exp}

To confirm that {\em trajectory servoing} outcomes translate to
practice, the short trajectory experiment is run on a LoCoBot equipped
with a RealSense stereo camera and an Intel NUC (i5-7260U).
Fig. \ref{fig:real_env} presents a top-view of the experimental
environment (left) and a robot view with SLAM features (right).
Based on the environment and how long features can be tracked within it,
the template trajectories are scaled down to $\sim$2.4m.
For pose-based control, two sources of robot pose estimates are
tested: robot odometry (RO) and SLAM.  Robot odometry is generated from
onboard wheel encoders and an IMU. RO will have imperfect odometry due
to measurement noise and uncertainty.  Five trials were run for each
trajectory and each tracking strategy. 
{Only terminal error is measured. The continuous
robot ground truth pose signal is unavailable.}

\subsubsection{Results and Analysis}
\label{sec:shortDist_real_exp_RA}

Table \ref{tab:shortTrajResults}(c) and Fig.~\ref{fig:shortTrajBench} give the
outcomes of the tested methods.  SLAM has the highest average terminal
error.  TS has the lowest TE average, being 42.9\% lower than RO and 56.0\%
lower than SLAM (p-values less than 1e-2 for both). 
The outcomes are consistent with those from simulation: 
trajectory servoing achieves better performance than pose-based tracking
strategies using V-SLAM over short segments. 

In one instance RO outperformed TS, the straight trajectory.  RO pose
estimation is more accurate for straight trajectories, such that odometry
drift is comparable to the image-based motion imperceptibility effects
that impact trajectory servoing performance under zero desired angular
velocities.  This observation will be seen again in real experiments
involving long distance trajectory servoing in the next section.

\subsubsection{Process Timing}
Image to control timing for trajectory servoing is $\sim\!\!26ms$.  The
major time cost is V-SLAM feature tracking, with tracked feature control
calculations taking $\sim\!\!1.5 ms$. V-SLAM image to pose estimate timing
is $\sim\!28 ms$.

%% file: longDist.tex
\newcommand{\trajTimeStart}[1]{t_{#1,s}}
\newcommand{\trajTimeEnd}[1]{t_{#1,e}}
\newcommand{\featureTraj}[1]{\featureSetD_{#1} (t) |_{\trajTimeStart{#1}}^{\trajTimeEnd{#1}}}

\section{Long Distance Trajectory Servoing}\label{sec:longDist}

Short distance trajectory servoing cannot extend to long trajectories due
to feature impoverishment. When moving beyond the initially visible
scene, a more comprehensive trajectory servoing system should augment
the feature pool $\featureSetD$ with new features. Likewise,
if navigation consists of multiple short distance trajectories, then
the system must have a regeneration mechanism for synthesizing entirely
new desired feature tracks for the new segment.  The overlapping needs
for these two events inform the creation of a module for feature
replenishment and trajectory extension.

\begin{figure*}[t]
\begin{minipage}[t]{0.3835\textwidth}
\centering
  \begin{tikzpicture}[inner sep=0pt, outer sep=0pt, auto, >=latex]
  \node[anchor=south west] (fig) at (0,0)
    	{{\scalebox{0.8}{\input{figs/trajservo_long.tex}}}};
  \end{tikzpicture}
  \caption{Feature replenishment process. There are three segments of feature trajectories. Stars are observed point sets at corresponding time. Each circle is the start or end time of next or this segment of feature trajectory. Three feature trajectories are generated by the feature replenishment equation \eqref{eq:trajReg}. \label{fig:feat_rep}}
\end{minipage}
\hfill
\begin{minipage}[t]{0.37\textwidth}
\vspace*{-6em}
\hspace*{-1em}
\centering
\tikzset{global scale/.style={
    scale=#1,
    every node/.append style={scale=#1}
  }
}
\begin{tikzpicture}[global scale = 0.5]
\node [draw,
    fill=Goldenrod,
    minimum width=1.5cm,
    minimum height=1cm,
    text width=2.5cm,
    text centered
]  (FG) at (0,0) {\centering \large Feature \\ Replenishment};

\node[draw,
    circle,
    minimum size=0.6cm,
    fill=Rhodamine!50,
    right=0.3cm of FG
] (sum) {};
 
\draw (sum.north east) -- (sum.south west)
    (sum.north west) -- (sum.south east);
 
\draw (sum.north east) -- (sum.south west)
(sum.north west) -- (sum.south east);
 
\node[left=-1pt] at (sum.center){\tiny $+$};
\node[below] at (sum.center){\tiny $-$};
 
\node [anchor=west, draw,
    fill=Goldenrod,
    minimum width=1.5cm,
    minimum height=1cm,
    text width=2.5cm,
    text centered,
    right=0.3cm of sum
]  (controller) {\centering \large Trajectory \\ Servoing Core};
 
\node [draw,
    fill=SpringGreen, 
    minimum width=1.2cm,
    minimum height=1cm,
    right=0.5cm of controller
] (system) {\centering \large Plant};
 
\node [draw,
    fill=SeaGreen, 
    minimum width=1.5cm,
    minimum height=1cm,
    text width=3cm,
    text centered,
    below right= 0.4cm and -0.8cm of controller
]  (sensor) {\centering \large V-SLAM \\ Tracked Features};

\node [draw,
    fill=SeaGreen, 
    minimum width=1.5cm,
    minimum height=1cm,
    below right= 1.2cm and -1.1cm of controller
]  (SLAM) {\centering \large V-SLAM Pose};
 
\draw[-stealth] (sum.east) -- (controller.west)
    node[midway,above,yshift=0.5em]{\resizebox{0.25cm}{!}{$\error$}};
 
\draw[-stealth] (controller.east) -- (system.west) 
    node[midway,above,yshift=0.5em]{\resizebox{0.9cm}{!}{$\dot{\featSetV},\roboVel$}};
    
\draw[-stealth] ($(controller.north) + (0in, 0.3in)$) -- (controller.north)
    node[midway,right]{\resizebox{0.5cm}{!}{$\roboVel^*$}};
 
\draw[-stealth] (system.east) -- ++ (1.25,0) 
    node[midway](output){}node[midway,above,yshift=0.5em,xshift=0.3em]{\resizebox{1.25cm}{!}{$\featSetV,\mcframe{\roboF}{W}{R}$}};
 
\draw[-stealth] (output.center) |- (sensor.east);
 
\draw[-stealth] (sensor.west) -| (sum.south) 
    node[near end,left]{\resizebox{0.3cm}{!}{$\hat{\featSetV}$}};
 
\draw [-stealth](FG) --(sum.west) 
    node[midway,above,yshift=0.5em]{\resizebox{0.5cm}{!}{$\featSetDV$}};
    
\draw [-stealth] ($(FG.west) + (-1.5,0)$) -- (FG.west)
    node[midway,above,yshift=0.5em,xshift=-0.3em]{\large Ref Traj};
    
\draw [-stealth] ($(FG.west) + (-1.5,0)$) -- (FG.west)
    node[midway,below,yshift=-0.3em,xshift=-0.3em]{\resizebox{1.1cm}{!}{$\mcframe{\roboF}{W,*}{R}$}};
    
\draw[-stealth] (output.center) |- (SLAM.east);
 
\draw[-stealth] (SLAM.west) -| (FG.south) 
    node[near end,left]{\resizebox{0.75cm}{!}{$\hat{{\roboF}}^{\mc{W}}_{\mc{R}}$}};
\end{tikzpicture}
\caption{Block diagram for long distance trajectory servoing. {Notations  follow \S\ref{sec:Intro} and \S\ref{sec:shortDist}.
$\featSetDV$, $\roboVel^*$ and $\hat{\featSetV}$, $\hat{{\roboF}}^{\mc{W}}_{\mc{R}}$ are the corresponding desired and measured values.} \label{fig:BlockTS}}

\end{minipage}
\hfill
\begin{minipage}[t]{0.22\textwidth}
\vspace*{-5em}
\centering
\hspace*{-0.5em}
  \begin{tikzpicture}[inner sep=0pt, outer sep=0pt, auto, >=latex]
  \node[anchor=south west] (fig) at (0,0)
    	{{\scalebox{0.6}{\input{figs/long_traj.tex}}}};
  \end{tikzpicture}
  \caption{Long distance trajectories. (a) Trajectories in simulation; (b) Trajectories in real experiments.\label{fig:long_trajs}}
\end{minipage}
\vspace*{-1em}
\end{figure*}

\subsection{Feature Replenishment}
\label{sec:longDist_FR}

The number of feature correspondences $\numFeat$ in $\featureSet$ and
$\featureSetD$ indicates whether trajectory servoing can be performed
without concern.  Let the threshold $\tau_{fr}$ determine when feature
replenishment should be triggered.
Define $\featureTraj{i}$ as the $i^{\rm th}$ feature trajectory starting from
$\trajTimeStart{i}$ and ending at $\trajTimeEnd{i}$. The case $i=0$ 
represents the first feature trajectory segment generated by
\eqref{eq:featDE} for $\trajTimeStart{i}=0$, integrated up to the 
maximum time $t_{\text{end}}$ of the given trajectory.
The time varying function $\numFeat(t)$ is the actual number of feature
correspondences between $\featureSet(t)$ and $\featureSetD_i (t)$ as
the robot proceeds.

When $\numFeat(t) \ge \tau_{fr}$, the feature trajectory
$\featureSetD_i(t)$ may be used for trajectory servoing. 
When $\numFeat(t) < \tau_{fr}$, the feature replenishment process will
be triggered at the current time and noted as $\trajTimeEnd{i}$. 
The old feature trajectory $\featureTraj{i}$ is finished. 
A new feature trajectory is generated with
\begin{equation} \label{eq:trajReg}
  \featureTraj{i+1} = \camProj \circ (\roboF^*(\trajTimeEnd{i},t)
  \mcframe{\camF}{R}{C})^{-1}(\pointSet^{\mc W}(\trajTimeEnd{i})),
\end{equation}
where $\roboF^*(\trajTimeEnd{i},t)$ is the transformation between
the current robot pose and a future desired pose ($t > \trajTimeEnd{i}$) on
the trajectory.  The poses behind the robot are not included. 
The set $\pointSet^{\mc W}(\trajTimeEnd{i})$ consists of observed
points at the current time $\trajTimeEnd{i}$. 
The feature pool is augmented by these current features.  When this
regeneration step is finished, the exact time will be assigned as the
$\trajTimeStart{i+1}$.  Trajectory servoing is performed on this new
feature trajectory until the regeneration is triggered again or the
arriving at the end of the trajectory.  The process of regenerating new
feature tracks is equivalent to dividing a long trajectory into a set
of shorter segments pertaining to the generated feature trajectory
segments. 
An depiction of feature replenishment is shown in Fig. \ref{fig:feat_rep}.

\begin{figure*}[t]
\vspace*{0.1in}
\begin{minipage}[t]{0.6\textwidth}
\captionof{table}{Long distance trajectory benchmark and real experiment results.\label{tb:longTrajResults}}
\begin{tikzpicture}[inner sep=0pt,outer sep=0pt,scale=1, every node/.style={scale=0.8}]

	\node[anchor=north west] (sim_ale) at (0, 0pt)
    {
    \setlength{\tabcolsep}{2pt}
    \begin{tabular}{|c||c|cc|c|}
    \hline 
    \textbf{Seq.} & PO & SLAM & TS & TS+PO \\ 
    \hline 
    LRU & 0.53  & 3.88  & 4.00  & 1.47 \\ 
    LLU & 0.86  & 8.21  & 5.18  & 1.61 \\ 
    LST & 1.13  & 5.03  & 3.00  & 2.03 \\ 
    LZZ & 1.06  & 7.54  & 5.90  & 5.04 \\ 
    \hline 
    \textbf{Avg.} & 0.90 & 6.17 & \textbf{4.52} & {2.54} \\
    \hline 
    \end{tabular}
    };
    
    \node[anchor=south, text centered] (sim_ale_cap) 
    at ($(sim_ale.north) + (0pt, 2pt)$)
    {\normalsize \textbf{(a)} Sim {ALE} (cm)};
      
    \node[anchor=west] (sim_te) at ($(sim_ale.east) + (2pt, 0pt)$)
    {
    \setlength{\tabcolsep}{2pt}
    \begin{tabular}{|c||c|cc|c|}
    \hline 
    \textbf{Seq.} & PO & SLAM & TS & TS+PO \\ 
    \hline 
    LRU & 8.57  & 10.66   & 6.42  & 4.15 \\ 
    LLU & 5.54  & 29.48  & 15.75 & 4.02 \\ 
    LST & 6.01  & 7.60  & 1.76  & 6.61 \\ 
    LZZ & 7.83  & 9.28   & 9.00  & 12.21 \\ 
    \hline 
    \textbf{Avg.} & 6.99 & 14.26 & \textbf{8.23} & 6.75 \\ 
    \hline 
    \end{tabular}
    };
    
	\node[anchor=south, text centered] (sim_te_cap) 
    at ($(sim_te.north) + (0pt, 2pt)$)
    {\normalsize \textbf{(b)} Sim Terminal Error (cm)};    
    
    \node[anchor=north west] (real_te) at ($(sim_te.north east) + (2pt, -10pt)$)
    {
    \setlength{\tabcolsep}{2pt}
    \begin{tabular}{|c||ccc|}
    \hline 
    \textbf{Seq.} & RO & SLAM & TS \\ 
    \hline 
    LS  & 8.2  & 14.1  & 10.9 \\ 
    LT & 12.8  & 11.8  & 6.8 \\ 
    \hline 
    \textbf{Avg.} & 10.5 & 13.0 & \textbf{8.9} \\ 
    \hline 
    \end{tabular}
    };
	
	\node[anchor=south, text width=3cm, text centered] (real_te_cap) 
    at ($(real_te.north) + (0pt, 2pt)$)
    {\normalsize \textbf{(c)} Real \\ Terminal Error (cm)};      

\end{tikzpicture}
  
\end{minipage}
\begin{minipage}[t]{0.4\textwidth}
\vspace*{0in}
\hspace*{-0.1in}
\centering
  \begin{tikzpicture}[inner sep=0pt, outer sep=0pt, auto, >=latex]
%

  \node[anchor=south west] (ale) at (0,0)
    	{{\scalebox{0.35}{\includegraphics[width=1.7in,clip=true,trim=0in 0in 0in 0in]{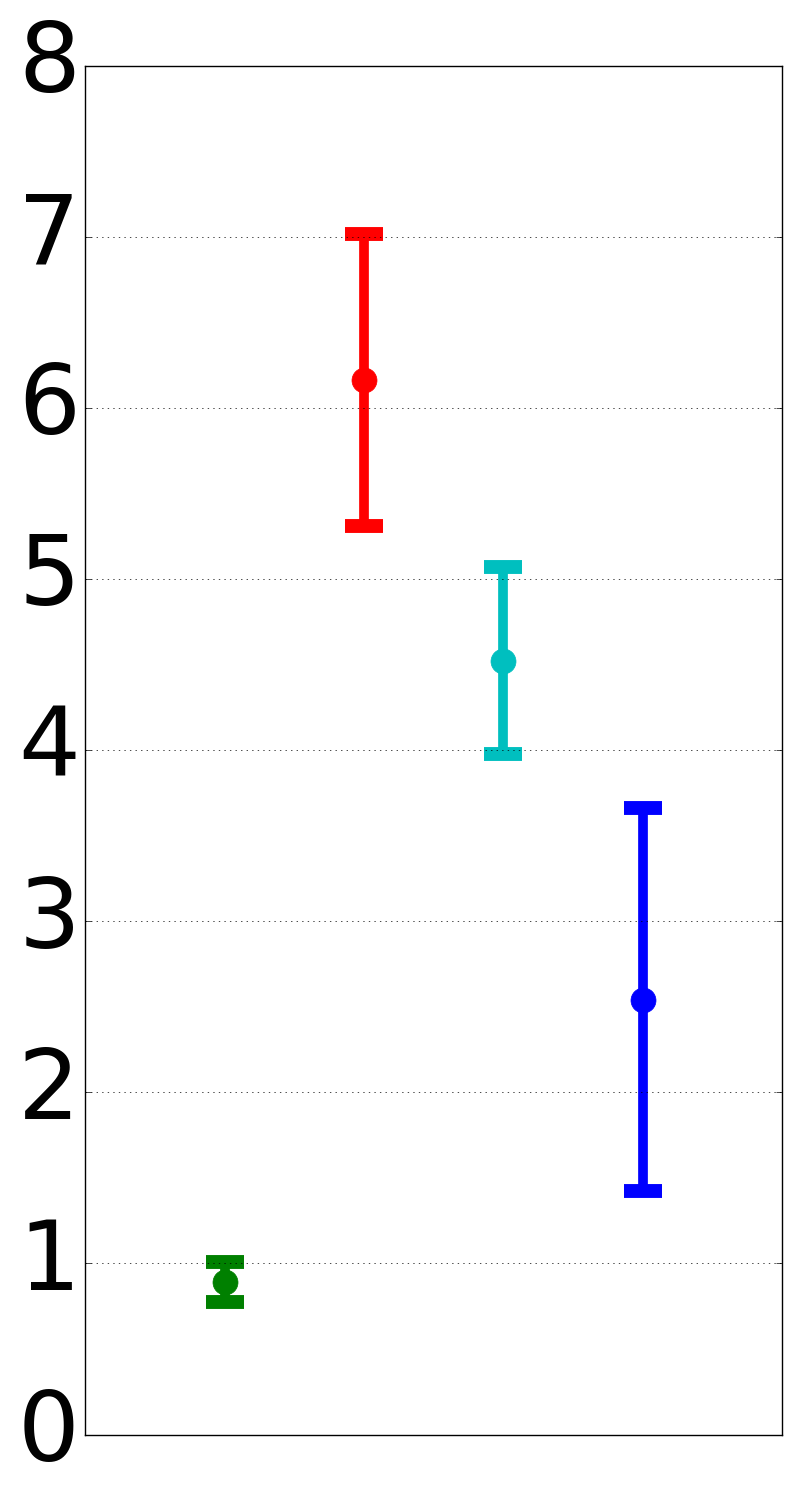}}}};
	
  \node[anchor=south, rotate=90] (col_yaxis_ale) at ($(ale.west) + (-2pt, 0)$) {\scriptsize Sim ALE (cm)};

  \node[anchor=west] (te) at ($(ale.east) + (12pt, 0)$)
    	{{\scalebox{0.35}{\includegraphics[width=1.7in,clip=true,trim=0in 0in 0in 0in]{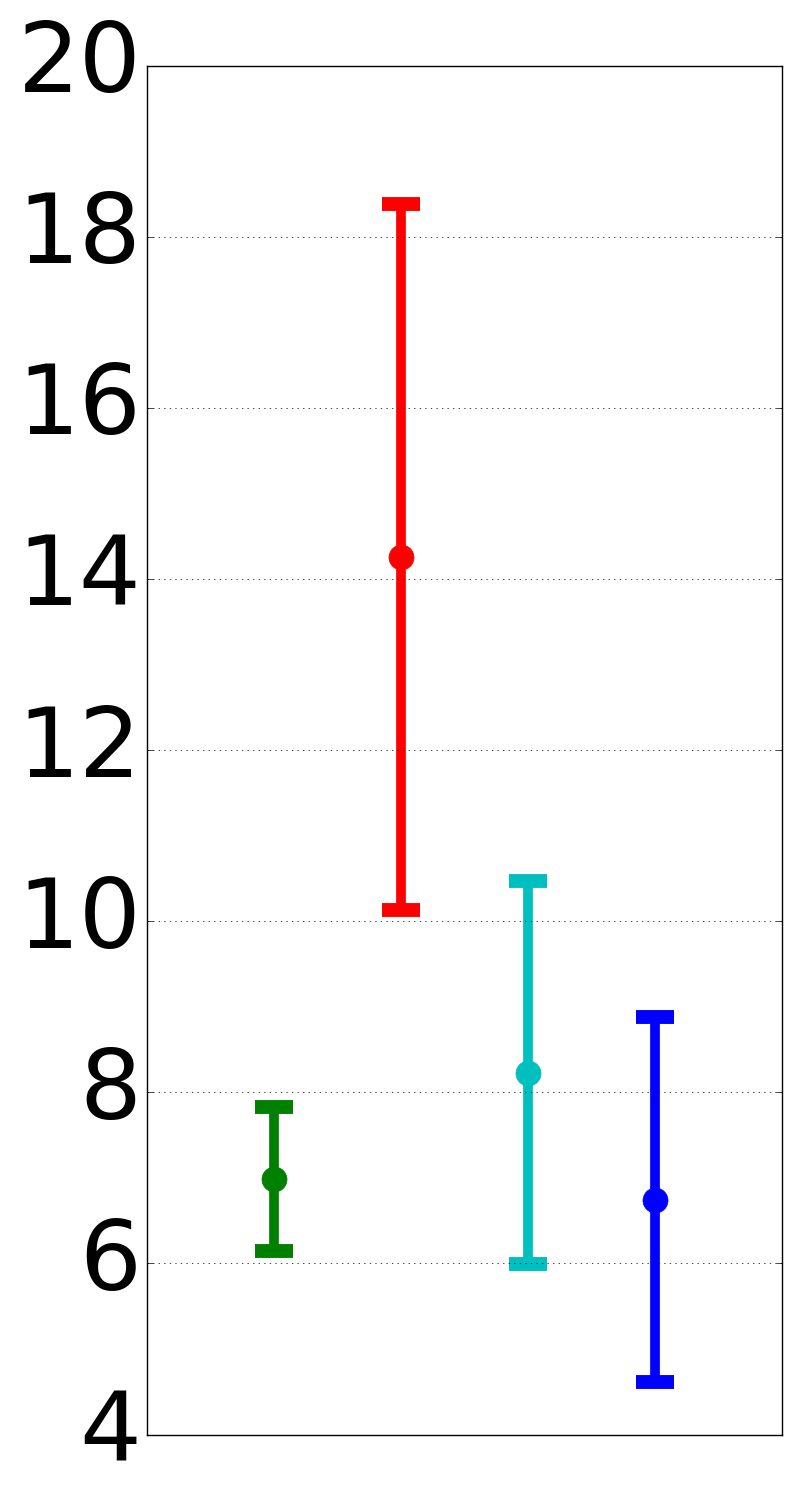}}}};
	
  \node[anchor=south, rotate=90] (col_yaxis_te) at ($(te.west) + (-2pt, 0)$) {\scriptsize Sim Terminal Error (cm)};
  
  \node[anchor=west] (rte) at ($(te.east) + (12pt, 0)$)
    	{{\scalebox{0.265}{\includegraphics[width=1.7in,clip=true,trim=0in 0in 0in 0in]{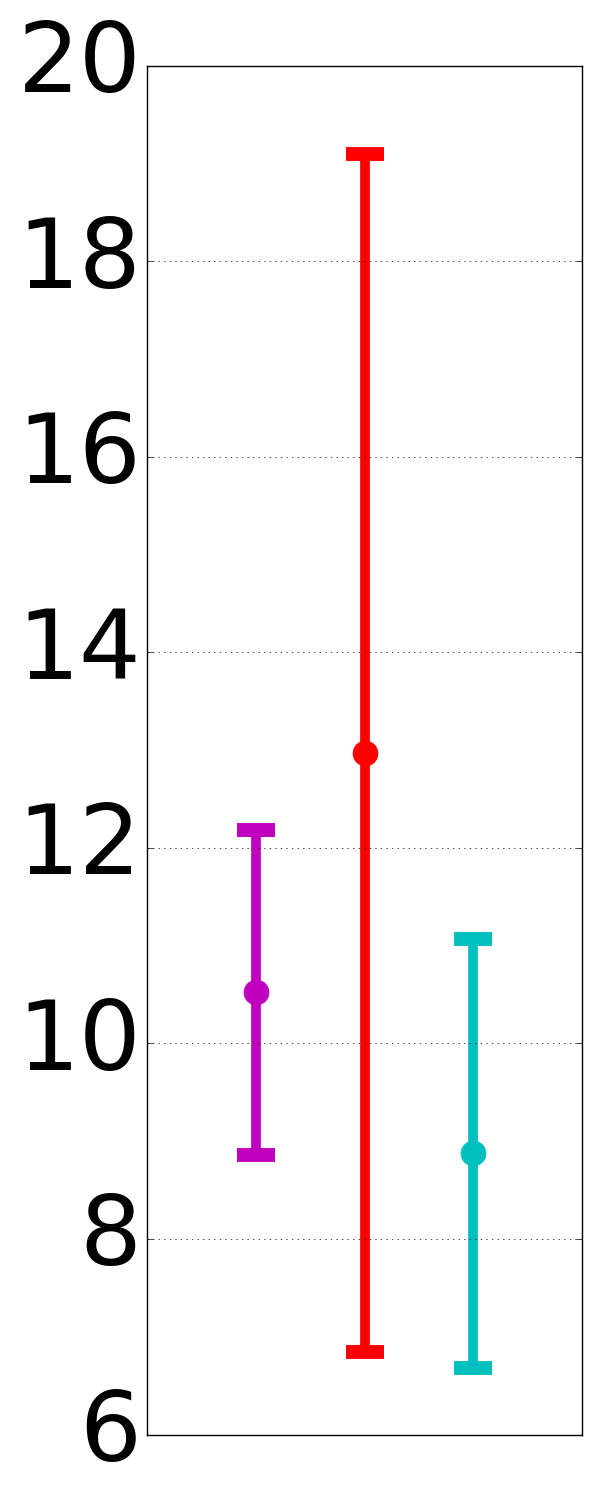}}}};
	
  \node[anchor=south, rotate=90] (col_yaxis_rte) at ($(rte.west) + (-2pt, 0)$) {\scriptsize Real Terminal Error (cm)};
  
  \node[anchor=west] (legend) at ($(rte.east) + (-1pt, 22pt)$)
    	{{\scalebox{0.25}{\includegraphics[width=1.7in,clip=true,trim=0in 0in 0in 0in]{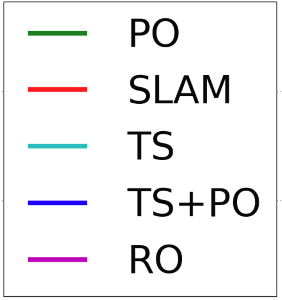}}}};
   
  \end{tikzpicture}
  \vspace*{-0.5em}
  \caption{95\% confidence intervals for long distance outcomes. 
  \label{fig:longTrajBench}}
  \vspace*{-1em}
\end{minipage}
\end{figure*}

During navigation, \eqref{eq:trajReg} requires the current robot pose
relative to the initial pose to be known.  In the absence of an
absolute reference or position measurement system, the only option
available is to use the estimated robot pose from V-SLAM, or some
equivalent process. 
Although there are some drawbacks to relying on V-SLAM, it attempts to
keep pose estimation as accurate as possible over long periods through
feature mapping, bundle adjustment, loop closure, etc. 
To further couple V-SLAM and trajectory servoing, we design a multi-loop
scheme, see Fig. \ref{fig:BlockTS}. 
The inner loop is governed by
trajectory servoing with V-SLAM tracked features. The V-SLAM estimated
pose is only explicitly used in the outer loop, during feature replenishment. 
{Though doing so may introduce bias into the feature trajectories, the
lower frequency reliance on SLAM avoids accumulating SLAM pose
estimation uncertainty as would higher frequency use in the inner loop
\cite{gitTS}.}

%% file: figs/trajservo_long.tex
\begin{tikzpicture}[inner sep=0pt, outer sep=0pt]
  \node[anchor=south west] (orig) at (0in,0in) 
    {{\includegraphics[width=0.8\columnwidth]{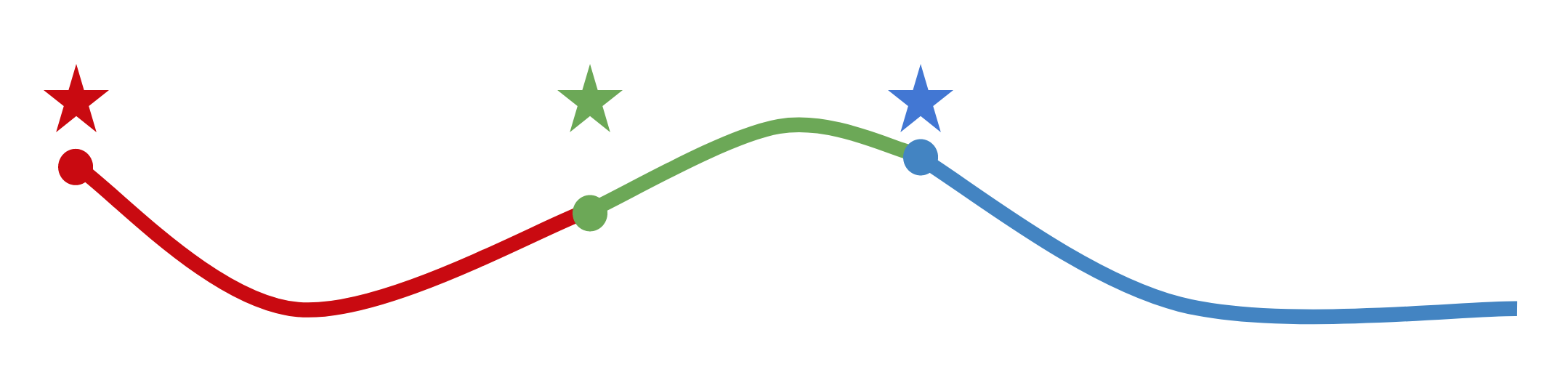}}};
  
  \node[anchor=west] (ti2_f) at ($(orig.west) + (-0.2in, 0.3in)$) {\footnotesize $\pointSet^{\mc W}(\trajTimeEnd{i-2})$};
  \node[anchor=west] (ti1_f) at ($(orig.west) + (0.5in, 0.3in)$) {\footnotesize $\pointSet^{\mc W}(\trajTimeEnd{i-1})$};
  \node[anchor=west] (ti_f) at ($(orig.west) + (1.2in, 0.3in)$) {\footnotesize $\pointSet^{\mc W}(\trajTimeEnd{i})$};
  
  \node[anchor=north] (ti21) at ($(ti2_f.south) + (0.05in, -0.4in)$) {\footnotesize $\trajTimeEnd{i-2}$};
  \node[anchor=north] (ti22) at ($(ti21.south) + (0in, 0in)$) {\footnotesize $\trajTimeStart{i-1}$};
  \node[anchor=north] (ti11) at ($(ti1_f.south) + (0.05in, -0.4in)$) {\footnotesize $\trajTimeEnd{i-1}$};
  \node[anchor=north] (ti12) at ($(ti11.south) + (0in, 0in)$) {\footnotesize $\trajTimeStart{i}$};
  \node[anchor=north] (ti1) at ($(ti_f.south) + (0in, -0.4in)$) {\footnotesize $\trajTimeEnd{i}$};
  \node[anchor=north] (ti2) at ($(ti1.south) + (0in, 0in)$) {\footnotesize $\trajTimeStart{i+1}$};
  
  \node[anchor=north] (tib_ft) at ($(orig.west) + (0.45in, -0.45in)$) {\footnotesize $\featureTraj{i-1}$};
  \node[anchor=north] (ti_ft) at ($(orig.west) + (1.13in, -0.45in)$) {\footnotesize $\featureTraj{i}$};
  \node[anchor=north] (tin_ft) at ($(orig.west) + (1.8in, -0.45in)$) {\footnotesize $\featureTraj{i+1}$};
  
  
  \draw[->, very thick] (tib_ft.north) -- ($(tib_ft.north) + (0in, 0.2in)$);
  \draw[->, very thick] (ti_ft.north) -- ($(ti_ft.north) + (0in, 0.5in)$);
  \draw[->, very thick] (tin_ft.north) -- ($(tin_ft.north) + (0in, 0.2in)$);
	
  \end{tikzpicture}

%% file: figs/long_traj.tex
\begin{tikzpicture}[inner sep=0pt, outer sep=0pt]
  \node[anchor=south west] (RU) at (0in,0in) 
    {{\includegraphics[scale=0.05,clip=true,trim=0in 0in 0.8in 0in]{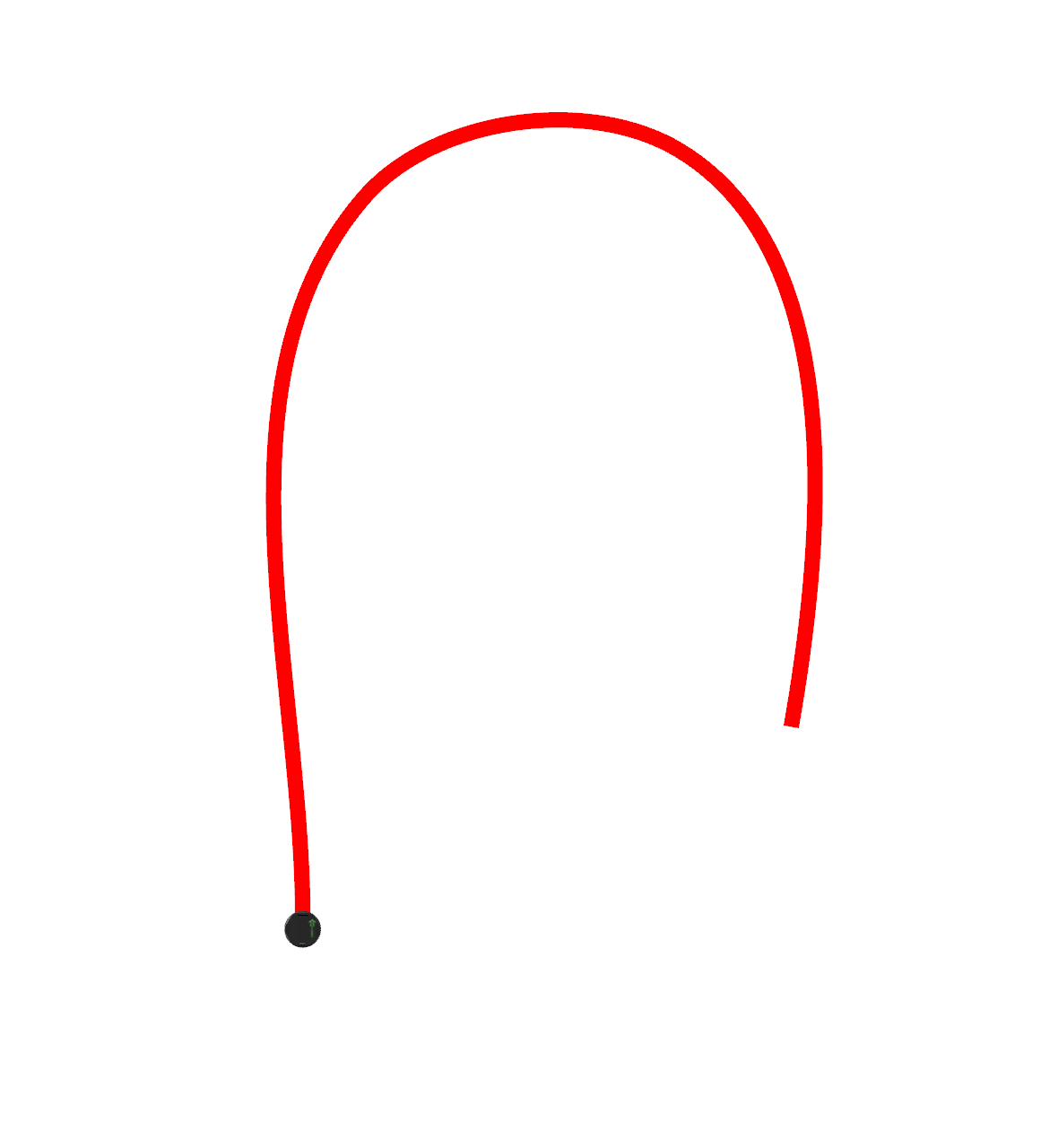}}};
  \node[anchor=south west,xshift=-10pt] (LU) at (RU.south east)
    {\includegraphics[scale=0.05,clip=true,trim=3in 0in 0in 0in]{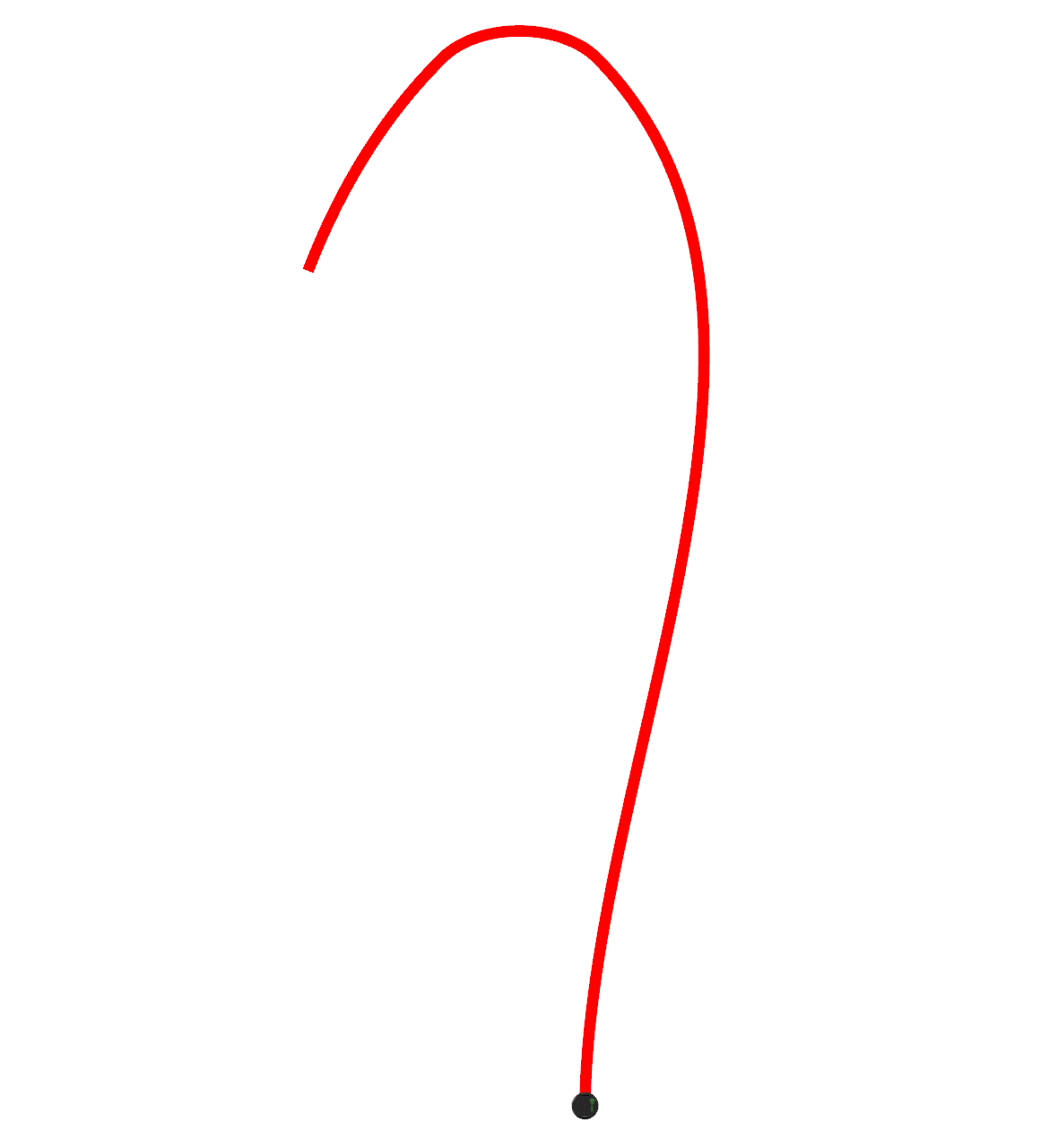}};
  \node[anchor=south west,xshift=-18pt] (ST) at (LU.south east)
    {\includegraphics[scale=0.05,clip=true,trim=2.5in 0in 0in 0in]{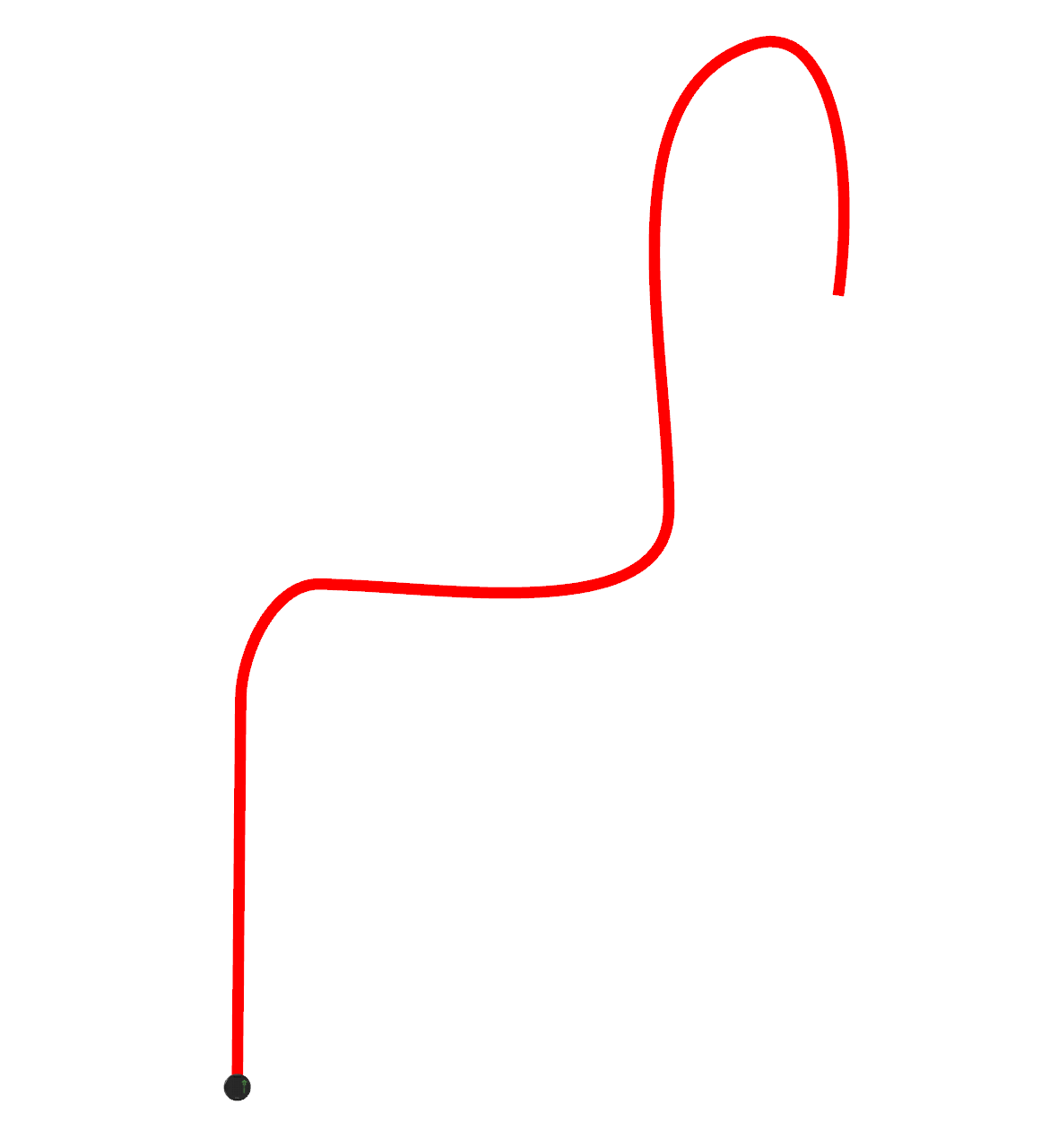}};
  \node[anchor=south west,xshift=-11pt] (ZZ) at (ST.south east)
    {\includegraphics[scale=0.05,clip=true,trim=2.5in 0in 0in 0in]{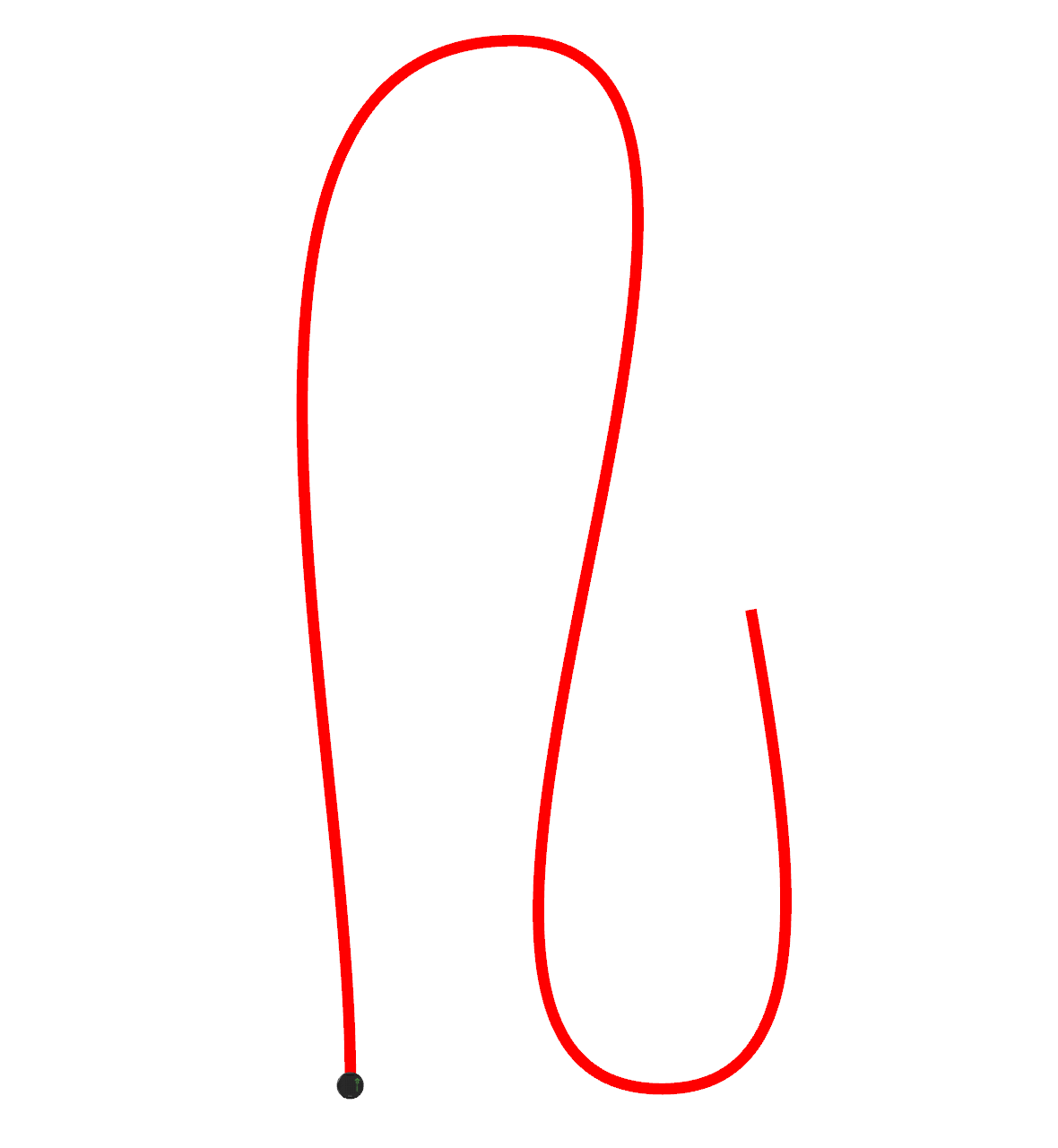}};
  \node[anchor=south west,xshift=-13pt] (LS) at (ZZ.south east)
    {\includegraphics[scale=0.05,clip=true,trim=1.2in 0in 0in 0in]{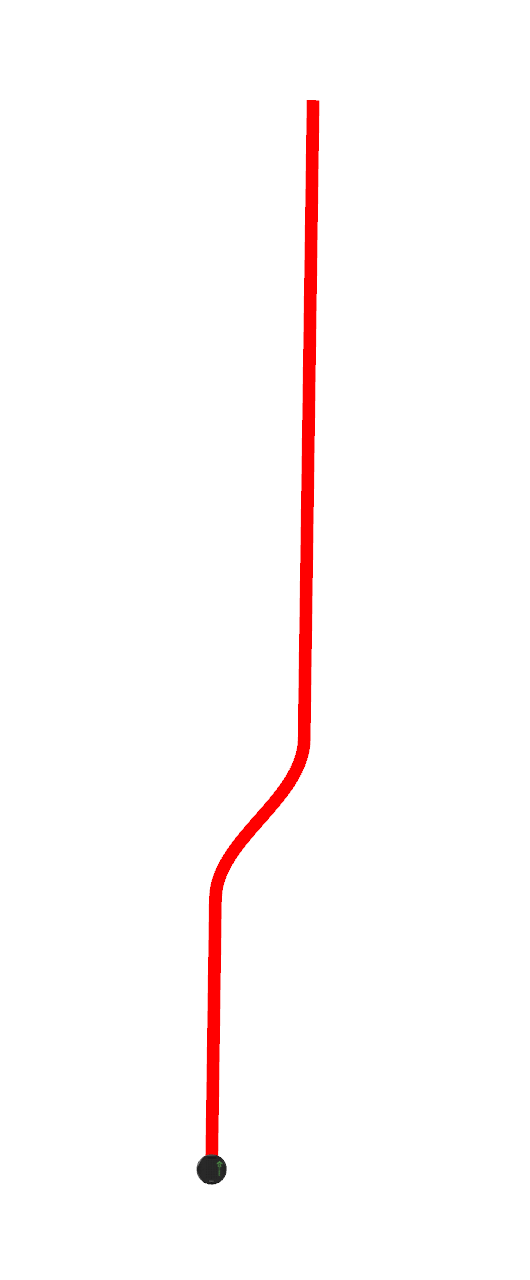}};
  \node[anchor=south west,xshift=-10pt] (LT) at (LS.south east)
    {\includegraphics[scale=0.03,clip=true,trim=1.2in 0in 0in 0in]{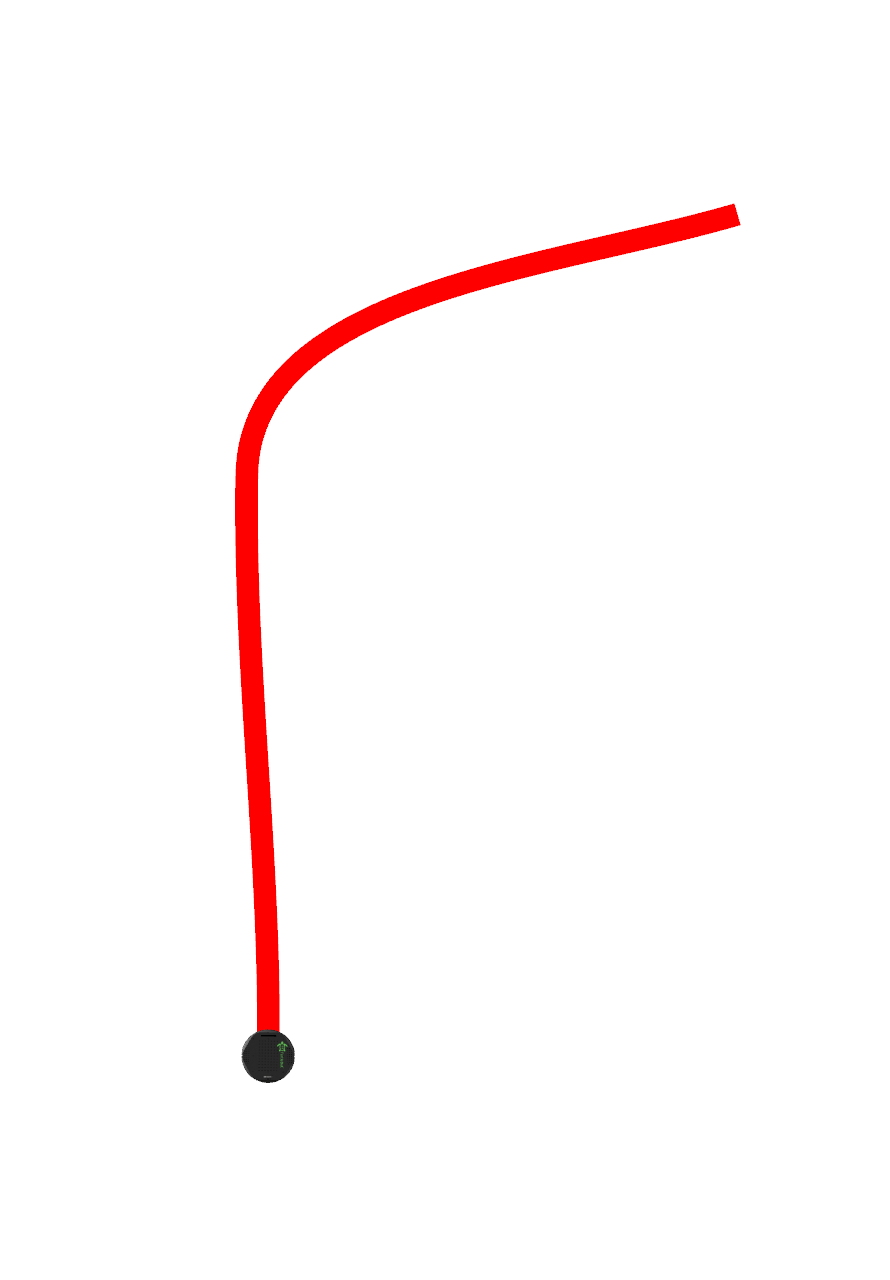}};

  \node[anchor=north] at ($(RU.south) + (2pt, -5pt)$) {\normalsize LRU};
  \node[anchor=north] at ($(LU.south) + (-5pt, -5pt)$) {\normalsize LLU};
  \node[anchor=north] at ($(ST.south) + (-5pt, -5pt)$) {\normalsize LST};
  \node[anchor=north] at ($(ZZ.south) + (-5pt, -5pt)$) {\normalsize LZZ};
  \node[anchor=north] at ($(LS.south) + (-4pt, -5pt)$) {\normalsize LS};
  \node[anchor=north] at ($(LT.south) + (0pt, -5pt)$) {\normalsize LT};
  
  \node[anchor=north west] at ($(RU.north west) + (5pt, 0)$) {\normalsize (a)};
  \node[anchor=north west] at ($(LS.north west) + (-5pt, 0)$) {\normalsize (b)};
  \end{tikzpicture}

%% file: longExp.tex
\subsection{Simulation Experiments and Results\label{sec:longDist_exp}}

This section modifies the experiments in \S\ref{sec:shortDist_exp} to
involve longer trajectories that trigger feature replenishment and
synthesize new feature trajectory segments. The set of trajectories to
track is depicted in Fig. \ref{fig:long_trajs}(a). They are denoted as long:
right u-turn (LRU), left u-turn (LLU), straight+turn (LST), and zig-zag
(LZZ). Each trajectory is around 20m or longer. 
Testing and evaluation follows as before (minus VS+).
A new tracking method is added: TS+PO, which uses perfect odometry
instead of SLAM odometry in the feature replenishment stage.
{The parameter was tuned for performance, giving 
$\tau_{fr}=10$ \cite{gitTS}.}

\subsubsection{Results and Analysis}
Tables \ref{tb:longTrajResults}(a,b) give outcomes for the two error metrics.
Though TS continues to outperform SLAM and is closer to PO, the gap
relative to PO is larger {than the gap for short trajectories.}
As hypothesized, longer trajectory error is affected by the need to use SLAM
pose estimates for regeneration.  SLAM pose drift impacts trajectory
tracking error for both methods, but is attenuated when using TS.  The TS+PO
outcomes confirm the impact of SLAM drift on TS performance as the TS+PO
outcomes are lower than the TS outcomes.
{Tighter coupling of the trajectory servoing and V-SLAM systems (e.g.,
using TS tracking to assist with pose estimation) should improve the
accuracy of pose estimation, and further improve tracking performance.}

{Fig. \ref{fig:longTrajBench} depicts the 95\% confidence
intervals of the outcomes for the template trajectories.} 
{For ALE, the intervals and p-values (\textless 5e-3) imply
that all pairwise comparisons are significant.} For TE, SLAM compared to
any method indicates statistical significance (p-values: \textless 8e-3).
The TE scores amongst PO, TS, and TS+PO are not significant 
(p-values: $>$0.16). TS and TS+PO performance is close to PO.

{For long distance trajectories, trajectory servoing again has
similar control effort to pose-based feedback.  However, the norms of the
time differentiated control signal are 0.151, 0.148, 0.565 and 0.518 for PO,
SLAM, TS and TS+PO respectively.  TS-based control signals remain less
smooth.} 

\subsection{Real Experiments and Results}\label{sec:longDist_real_exp}

The last experiment evaluates long trajectory performance on a real
robot.  The experimental setup is similar to \S\ref{sec:shortDist_real_exp}. 
Two long trajectories, LS and LT, in Fig. \ref{fig:long_trajs}(b) are
used, of lengths $\sim$13m and $\sim$8m, respectively.

\subsubsection{Results and Analysis}

In Table \ref{tb:longTrajResults}(c), TS ranks first for the average 
TE metric. Average TE decreases 31.5\% from SLAM to TS and 15.2\% from RO to
TS. Importantly TS outperforms SLAM in all cases, with lower variance
(see Fig.~\ref{fig:longTrajBench}).
These outcomes are consistent with the simulation results and support
the premise behind the trajectory servoing system design described at
the end of \S \ref{sec:longDist_FR}.

{For the LS trajectory with long straight segments, RO outperforms SLAM 
because it has better pose estimation. 
Using TS reduces the terminal error by 22.7\%, compared to SLAM. 
Trajectory servoing is more robust to pose estimation errors. 
Yet, it is still impacted by the SLAM pose uncertainty arising from
forward motion imperceptibility due to TS' use of SLAM state information
for feature regeneration.
These results reproduce the observations in \S \ref{sec:shortDist_real_exp_RA}.}

\subsubsection{Feature Replenishment Frequency}
Trajectory servoing feature replenishment calls queried SLAM poses 1.5 times per
meter in simulation and experiment.
The SLAM pose-based tracking method equivalent is 100 times per meter.

%% file: conc.tex
\section{Conclusion \label{sec:con}}

The paper presented an image-based trajectory tracking approach for a
non-holonomic mobile ground robot. Called {\em trajectory servoing},
it combines IBVS and V-SLAM to achieve tracking through unknown
environments without externally derived absolute positioning information.
Trajectory servoing successfully follows short trajectories.
Using estimated robot poses from the V-SLAM module extends trajectory
tracking to longer trajectories.  
Experiments demonstrate improved accuracy over pose-based trajectory
tracking using estimated SLAM poses or robot odometry.
Trajectory servoing is less impacted by pose
estimation error, by virtue of directly using for feedback the features
from which pose is inferred.
However, the trade-off for relying on visual features is
that feature poor environments may not be traversable using trajectory
servoing.

Real-world uncertainty and disturbances as well as non-smooth
trajectories may degrade tracking performance. 
Future work intends to improve robustness of trajectory servoing 
by adding linear velocity control, temporal smoothness constraints, and
performing uncertainty analysis on the feedback equations.  In addition,
more tightly coupling trajectory servoing with V-SLAM may benefit the
pose estimation process and further improve tracking performance.

%% file: append.tex
\section{Study of using pose with higher frequency}

In this study, we are investigating the effect that pose estimation 
uncertainty brings to tracking performance. 
In SLAM pose-based control, the pose estimation errors immediately cause worse tracking performance. 
The design of trajectory servoing intends to improve the performance 
by less frequently requesting robot poses and shifting to the image domain. 
We already have the comparison of SLAM and TS in the paper. 
To further show the effect of pose estimation uncertainty, 
we conduct ablation studies about tracking performance 
when trajectory servoing more frequently using SLAM poses.

\subsection{Feature replenishment threshold $\tau_{fr}$ \label{sec:thresh}}

\begin{figure}[ht]
\begin{minipage}[t]{0.49\textwidth}
\centering
\begin{tikzpicture}[inner sep=0pt,outer sep=0pt]
	  \node[anchor=south west] (pop) at (0in, 0in)
      {{\includegraphics[height=1.8in,clip=true,trim=0in 0in
      0in 0in]{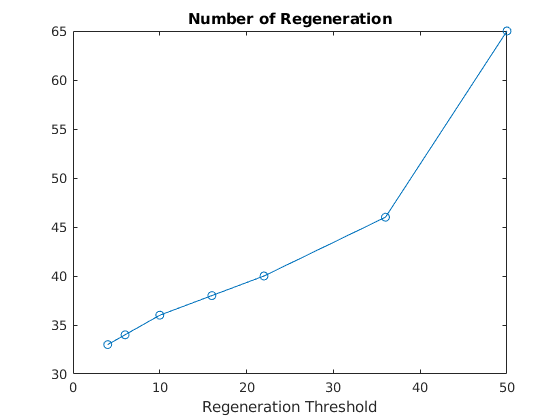}}};
\end{tikzpicture}
\end{minipage}
\hfill
\begin{minipage}[t]{0.49\textwidth}
\centering
\begin{tikzpicture}[inner sep=0pt,outer sep=0pt]
	  \node[anchor=south west] (pop) at (0in, 0in)
      {{\includegraphics[height=1.8in,clip=true,trim=0in 0in
      0in 0in]{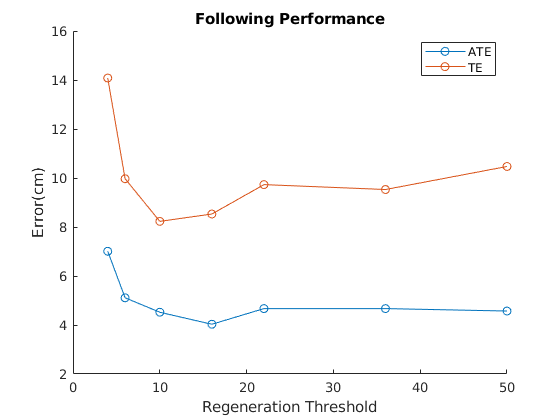}}};
\end{tikzpicture}
\end{minipage}
\vspace*{-0.5em}
\caption{Ablation study plots of feature replenishment $\tau_{fr}$ \label{fig:regThresh}}
\vspace*{-0.5em}
\end{figure}

From \S III.A in the paper, the feature replenishment threshold $\tau_{fr}$ 
controls the frequency of feature trajectory regeneration and 
robot pose requesting with long distance trajectories.
In the simulation, we conduct an ablation study about the influence of 
varying $\tau_{fr}$. 
The same two trajectory tracking performance metrics are used: 
average lateral error (ALE) and terminal error (TE).
The number of feature trajectory regeneration per length is also recorded.
Outcomes of averages over all trajectory templates are included in 
Table \ref{tab:regThresh} and Fig. \ref{fig:regThresh}

With smaller $\tau_{fr}$, trajectory servoing uses less number of tracked features, 
and works similar to pure IBVS. 
The computed control could be unstable and have oscillations, which reduces 
the tracking performance. 
In contrast, if $\tau_{fr}$ is significantly large, trajectory servoing 
will more frequently trigger feature replenishment. 
The performance will be more related to pose estimation accuracy, 
and be close to SLAM pose-based control.
More pose estimation errors cause worse performance.

\begin{figure}[ht]
\begin{minipage}[ht]{\columnwidth}
\centering
\captionof{table}{ Raw data from ablation study of feature replenishment $\tau_{fr}$ \label{tab:regThresh}}
\vspace*{-0.5em}
\begin{tikzpicture}[inner sep=0pt,outer sep=0pt,scale=1, every node/.style={scale=0.8}]
  \node[anchor=north west] (sim_thresh) at (0, 0pt)
  {
  \setlength{\tabcolsep}{4pt}
  \begin{tabular}{|c||ccccccc|}
  \hline 
  \textbf{$\tau_{fr}$} & 4 & 6 & \textbf{10} & 16 & 22 & 36 & 50 \\ 
  \hline 
  \textbf{ALE}  & 7.01 & 5.11 & 4.52 & \textbf{4.03} & 4.67 & 4.67 & 4.57 \\ 
  \textbf{TE}   & 14.08 & 9.97 & \textbf{8.23} & 8.53 & 9.73 & 9.53 & 10.47  \\ 
  \textbf{\# of Reg/m} & \textbf{1.4} & 1.5 & 1.6 & 1.7 & 1.8 & 2.0 & 2.7 \\ 
  \hline 
  \end{tabular}
  };       

\end{tikzpicture}
\end{minipage}
\vspace*{-0.5em}
\end{figure}

\subsection{Incremental Trajectory Servoing}

In order to further investigate the effect that pose estimation uncertainty 
brings to tracking performance. 
The trajectory servoing system is modified to use robot pose information 
in every control loop and recomputes the desired feature set 
for the next pose on the given trajectory. 
The feature trajectory will not be precomputed at the beginning.
This version is called {\em Incremental Trajectory Servoing} (I-TS).
Unlike the regular feature replenishment in \S III.A that is triggered 
by $\tau_{fr}$, desired feature regeneration is triggered by time.
By doing this modification, trajectory servoing works similar to 
SLAM pose-based control, but shifting feedback to the image domain. 

We applied the same simulation benchmark and metrics (i.e. ALE and TE) 
to compare tracking performance. 
The benchmark is only tested with long distance trajectories, 
because SLAM pose estimation uncertainty is not explicit with short distance 
trajectories. Tables \ref{tab:longITS}(a,b) gives outcomes for the two metrics.
I-TS has worse ALE and TE than TS and is similar to SLAM.
It shows that frequently using SLAM pose information will introduce more 
uncertainty that potentially do harm to the tracking.

\begin{figure}[ht]
\begin{minipage}[t]{\columnwidth}
\centering
\captionof{table}{ Incremental Trajectory Servoing Outcomes.\label{tab:longITS}}
\vspace*{-0.5em}
\begin{tikzpicture}[inner sep=0pt,outer sep=0pt,scale=1, every node/.style={scale=0.8}]
	\node[anchor=north west] (sim_ale) at (0, 0pt)
    {
    \setlength{\tabcolsep}{4pt}
    \begin{tabular}{|c||c|ccc|}
    \hline 
    \textbf{Seq.} & PO & SLAM & TS & I-TS \\ 
    \hline 
    LRU & 0.53  & 3.88  & 4.00  & 3.56 \\ 
    LLU & 0.86  & 8.21  & 5.18  & 7.68 \\ 
    LST & 1.13  & 5.03  & 3.00  & 4.19 \\ 
    LZZ & 1.06  & 7.54  & 5.90  & 9.72 \\ 
    \hline 
    \textbf{Avg.} & 0.90 & 6.17 & \textbf{4.52} & 6.29 \\
    \hline 
    \end{tabular}
    };
    
    \node[anchor=south, text width=5cm, text centered] (sim_ale_cap) 
    at ($(sim_ale.north) + (0pt, 2pt)$)
    {\normalsize \textbf{(a)} Sim {ALE} (cm)};
      
    \node[anchor=west] (sim_te) at ($(sim_ale.east) + (5pt, 0)$)
    {
    \setlength{\tabcolsep}{4pt}
    \begin{tabular}{|c||c|ccc|}
    \hline 
    \textbf{Seq.} & PO & SLAM & TS & I-TS \\ 
    \hline 
    LRU & 8.57  & 10.66   & 6.42  & 8.82 \\ 
    LLU & 5.54  & 29.48  & 15.75 & 24.3 \\ 
    LST & 6.01  & 7.60  & 1.76  & 10.45 \\ 
    LZZ & 7.83  & 9.28   & 9.00  & 9.91 \\ 
    \hline 
    \textbf{Avg.} & 6.99 & 14.26 & \textbf{8.23} & 13.37 \\ 
    \hline 
    \end{tabular}
    };
    
	\node[anchor=south, text width=5cm, text centered] (sim_ale_p_cap) 
    at ($(sim_te.north) + (0pt, 2pt)$)
    {\normalsize \textbf{(b)} Sim {TE} (cm)};        

\end{tikzpicture}
\end{minipage}
\vspace*{-1.5em}
\end{figure}

\subsection{Conclusion}

The above two studies shows the effect of pose estimation uncertainty 
when using SLAM poses with higher frequency.
In addition, the results of TS+PO in the paper also 
indicate the improved performance with more accurate pose information. 
Overall, pose estimation uncertainty can lead to worse trajectory tracking performance. 
Our trajectory servoing design can reduce the effect of this uncertainty 
and have better tracking results.